\documentclass{article}

\PassOptionsToPackage{numbers, compress}{natbib}

\usepackage[preprint]{neurips_2021}

\usepackage[utf8]{inputenc} 
\usepackage[T1]{fontenc}    
\usepackage[colorlinks=true,linkcolor=blue,citecolor=blue]{hyperref}       
\usepackage{url}            
\usepackage{booktabs}       
\usepackage{amsfonts}       
\usepackage{nicefrac}       
\usepackage{microtype}      
\usepackage[table,xcdraw]{xcolor}
\usepackage[most]{tcolorbox}
\usepackage{enumitem}
\usepackage{pifont}

\usepackage{amsmath}
\usepackage{amssymb}
\usepackage{mathtools}
\usepackage{amsthm}
\usepackage{graphicx}
\usepackage{subfigure}
\usepackage{caption}
\usepackage{tabularx}
\theoremstyle{plain}

\theoremstyle{definition}

\theoremstyle{remark}

\usepackage[capitalize,noabbrev]{cleveref}
\usepackage{framed}
\usepackage{multirow}
\usepackage{svg}
\usepackage{listings}
\usepackage{wrapfig}

\usepackage{amsmath,amsfonts,bm}

\def\eqref#1{equation~\ref{#1}}

\def\1{\bm{1}}

\DeclareMathAlphabet{\mathsfit}{\encodingdefault}{\sfdefault}{m}{sl}
\SetMathAlphabet{\mathsfit}{bold}{\encodingdefault}{\sfdefault}{bx}{n}

\newcommand\ours{\textsc{MetaLM}}
\Crefname{Section}{Section}{Section}

\newcommand {\otoprule }{\midrule [\heavyrulewidth ]}

\title{Language Models are General-Purpose Interfaces}

\author{Yaru Hao\thanks{~Equal contribution. $\dagger$ Corresponding author.}, ~~Haoyu Song\footnotemark[1], ~~Li Dong\footnotemark[1] \\
\bf Shaohan Huang, ~~Zewen Chi, ~~Wenhui Wang, ~~Shuming Ma, ~~Furu Wei$^\dagger$ \\
Microsoft Research \\
\url{https://github.com/microsoft/unilm}
}

\begin{document}

\maketitle

\begin{abstract}
Foundation models have received much attention due to their effectiveness across a broad range of downstream applications.
Though there is a big convergence in terms of architecture, most pretrained models are typically still developed for specific tasks or modalities.
In this work, we propose to use language models as a general-purpose interface to various foundation models.
A collection of pretrained encoders perceive diverse modalities (such as vision, and language), and they dock with a language model that plays the role of a universal task layer.
We propose a semi-causal language modeling objective to jointly pretrain the interface and the modular encoders.
We subsume the advantages and capabilities from both causal and non-causal modeling, thereby combining the best of two worlds.
Specifically, the proposed method not only inherits the capabilities of in-context learning and open-ended generation from causal language modeling, but also is conducive to finetuning because of the bidirectional encoders.
More importantly, our approach seamlessly unlocks the combinations of the above capabilities, e.g., enabling in-context learning or instruction following with finetuned encoders.
Experimental results across various language-only and vision-language benchmarks show that our model outperforms or is competitive with specialized models on finetuning, zero-shot generalization, and few-shot learning.
\end{abstract}

\begin{figure*}[h]
\centering
\includegraphics[width=0.82\columnwidth]{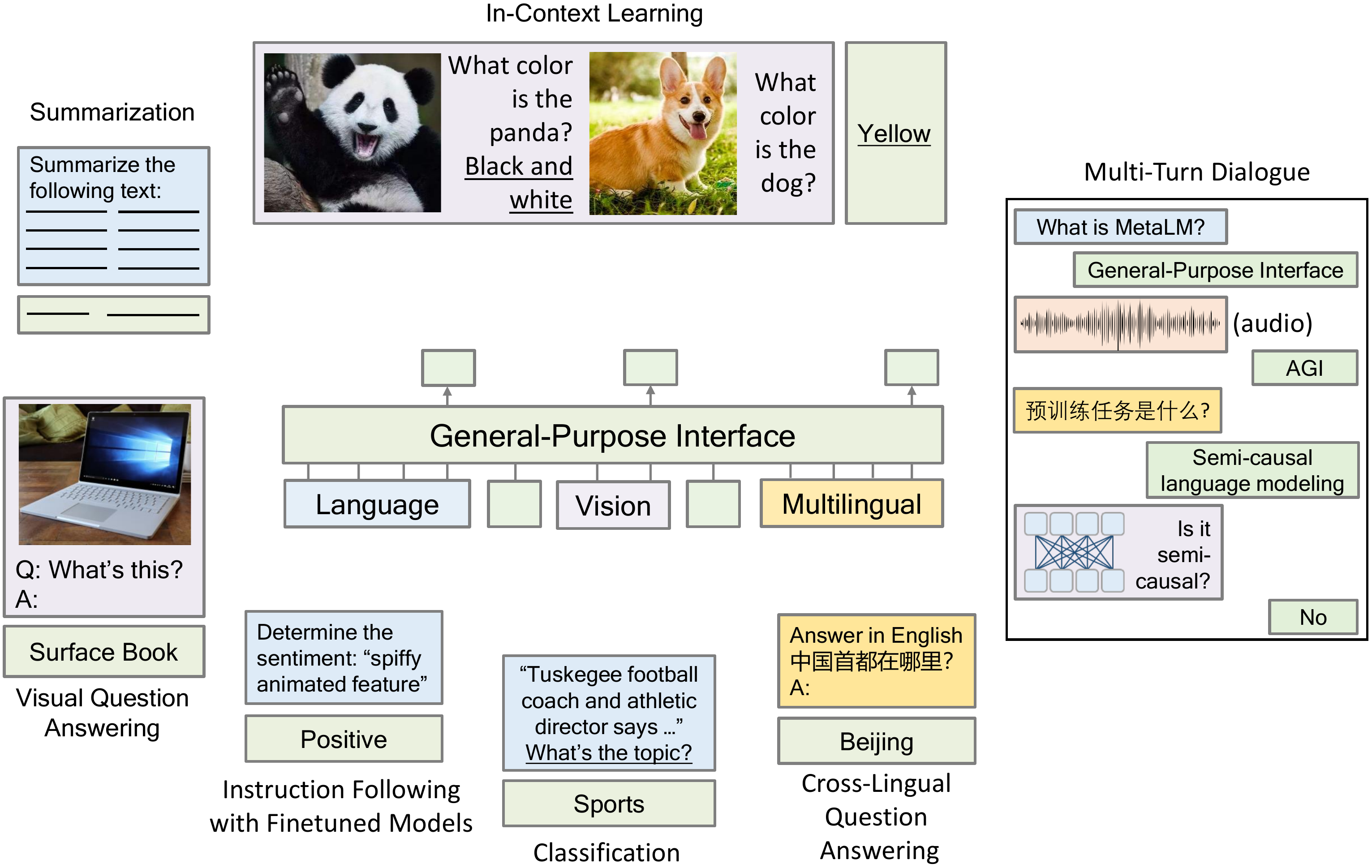}
\caption{Language models as a general-purpose interface to various foundation models.}
\label{fig:capability}
\end{figure*}

\newpage

\tableofcontents

\newpage

\section{Introduction: Design Principles}
\label{sec:intro}

\paragraph{Language models as a universal task layer.}
The large-scale language model serves as a general-purpose interface not only for language tasks, but also for vision, and multimodal tasks.
Language models have open-ended output space, which generalizes to a wide range of tasks.
As long as we can describe the predictions via natural language, the downstream task can fit in with language-model-based task layer.
It is natural that transforming various predictions to free-text sequences~\citep{t5}.
For example, we can transform the target labels, and answers to texts for classification, and question answering, respectively.
In addition, with the help of the universal task layer, the prediction process can go beyond single turn, i.e., a multi-turn dialogue interface can be built upon language models by conditioning on history context.
Such unification of various tasks is important to general-purposed AI, which unifies representations, transformations, and expressions into a shared module.

\paragraph{Causal language modeling (i.e., unidirectional decoder) is conducive to zero-shot generalization and in-context learning.}
GPT-3~\citep{gpt3} has shown that the intriguing properties emerge from causal language model pretraining.
Because of the favorable sample efficiency and inductive bias~\citep{what:lm:objective} of causal language modeling (i.e., all tokens make predictions and produce supervision signals) compared with other counterparts (such as masked language modeling), it is effective to give models the desired properties via causal language modeling.
The capabilities of zero- and few-shot learning are critical to be a general-purpose task layer.
Zero-shot generalization indicates that language models have learned an enormous amount of world knowledge~\citep{knowledgeneurons} and patterns by reading large-scale text corpora.
The memorized information can serve as reusable background knowledge and basic skills for a wide range of end tasks.
Moreover, in-context learning enables us to easily adapt either pretrained or finetuned models to new scenarios.
For example, we can use task instructions~\citep{instructgpt} to repurpose the model, and use demonstrations of some examples to conduct few-shot learning.

\paragraph{Non-causal modeling (i.e., bidirectional encoder) is conducive to transfer across tasks, languages, and modalities.}
Although causal language models are good at zero- and few-shot generalization, BERT~\citep{bert} and T5~\citep{t5} show that having bidirectional encoders pretrained by masked language modeling achieves much better finetuning performance.
Once the whole input is given, non-causal modeling is quite rational for encoding data.
Because all the context can access each other, while causal modeling can only make use of history tokens one by one.
The advantage of finetuning is helpful for the data-rich setting where there are many annotated data available.
In addition, non-causal encoder pretrained by the masked language modeling objective achieves competitive performance on cross-lingual transfer~\citep{xlmr}, which makes it effective to adapt models to the multilingual setting.

\paragraph{Semi-causal language modeling as a meta-pretraining task.}
Semi-causal language modeling plays the role of linking together non-causal encoders and the causal language model.
It is a meta task in the sense of universal interface pretraining of pretrained encoders.
Specifically, non-causal encoders learn to represent various input data, and a causal language model serves as a universal task layer.
Non-causal encoders dock with a causal language model, so that we can benefit from both modeling methods described as above.
In comparison with previous encoder-decoder pretraining (such as prefix language modeling, and T5;~\citealt{t5}), our task non-causally encodes random spans of the whole sequence, while generating the rest via causal language modeling.
Moreover, in terms of architecture, we directly feed the outputs of bidirectional encoders into the causal decoder, rather than relying on cross attention~\citep{transformer}.
Besides, multiple bidirectional encoders can be mounted to the causal language model, but the encoder-decoder architecture usually has only one encoder.

\paragraph{Non-causal encoders as System 1, and causal language models as System 2.}
Cognition is usually categorized into two levels~\citep{think:fast:slow,system12bengio}: System 1 (i.e., intuitive, and unconscious) and System 2 (i.e., sequential, conscious, planning, and reasoning).
In the proposed framework, the modules can be regarded as an implementation of these two levels, respectively.
To be specific, non-causal encoders pretrained by masked data modeling, such as BERT~\citep{bert} and BEiT~\citep{beit}, are used as a perception layer to encode various input modalities.
The encoding modules can be viewed as System 1.
After we obtain the input representations, we feed them to the causal language model, which has shown promising performance on commonsense reasoning~\citep{palm} and planning~\citep{gpt:planner}.
The universal task layer is designed to play a role of System 2 in our method.

\paragraph{Natural language interface between users and pretrained models.}
The universal task layer based on causal language modeling enables users to interact with pretrained non-causal encoders using natural language.
First, language can be used as a programming language for the underlying pretrained or finetuned models, which is compiled by the universal interface.
For example, we can write text-based instructions~\citep{instructgpt} and explanations~\citep{chain:of:thought} to repurpose and guide the model behaviors.
Second, the universal interface enables the models to present the results using free texts, making predictions directly understandable and explainable.
Third, the proposed framework natively supports multi-turn conversational interactions.
In each turn, we can feed the encoded input to the interface layer and then generate response results in a semi-causal manner.

\section{\ours{}: Meta Language Model}
\label{sec:method}

\begin{figure*}[t]
\centering
\includegraphics[width=\columnwidth]{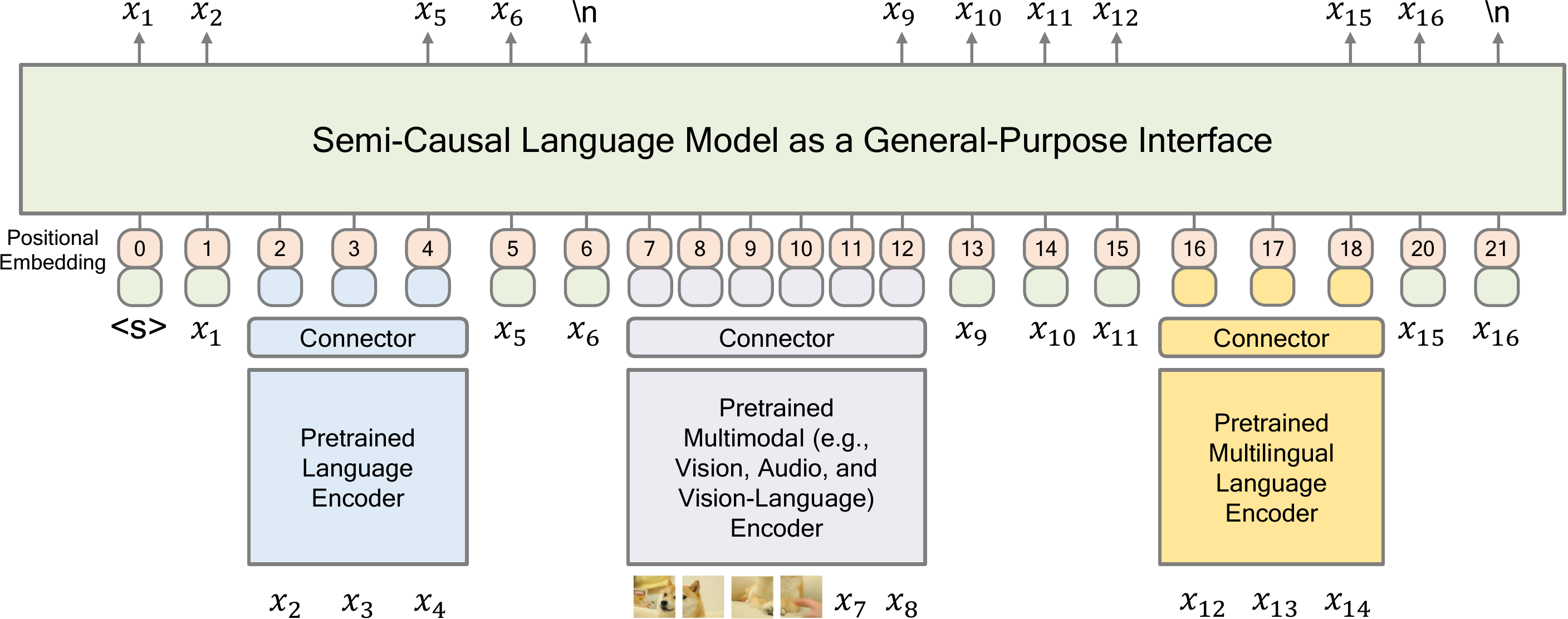}
\caption{Overview of \ours{}. The semi-causal language model serves as a general-purpose interface and supports interactions with various foundation models.}
\label{fig:overview}
\end{figure*}

Guided by the design principles in Section~\ref{sec:intro}, we present \textbf{Meta} \textbf{L}anguage \textbf{M}odel (\ours{}), a semi-causal language model that plays the role of a general-purpose interface and supports interactions with various foundation models.
An overview of our framework is shown in Figure~\ref{fig:overview}.
Specifically, a collection of pretrained encoders, that perceive diverse modalities, dock with a language model.
The language model is regarded as a universal task layer (i.e., general-purpose interface), which unifies various tasks as free-text generation.

In order to pretrain \ours{}, we propose a semi-causal language modeling task to jointly learn the modules.
\ours{} subsumes the advantages and capabilities from both worlds.
From the language model, \ours{} inherits the capabilities of in-context learning, multi-turn interaction, and open-ended generation.
Moreover, the underlying foundation models are conducive to finetuning because of bidirectional modeling~\citep{what:lm:objective}.

\subsection{Input Representation}

Input representations of \ours{} are grouped into two categories.
The first type is contextualized representations obtained by the underlying encoders and then projected by a connector layer.
For example, as shown in in Figure~\ref{fig:overview}, the image patches and $x_7, x_8$ are encoded by the bidirectional vision-language encoder.
The second category is token embeddings of texts, such as $x_5$, and $x_6$ in Figure~\ref{fig:overview}.
The representations of these two categories are summed with positional embeddings before feeding into the general-purpose interface.

\subsection{Model Architecture}
\label{sec:method:arch}

\begin{figure*}[t]
\centering
\includegraphics[width=\columnwidth]{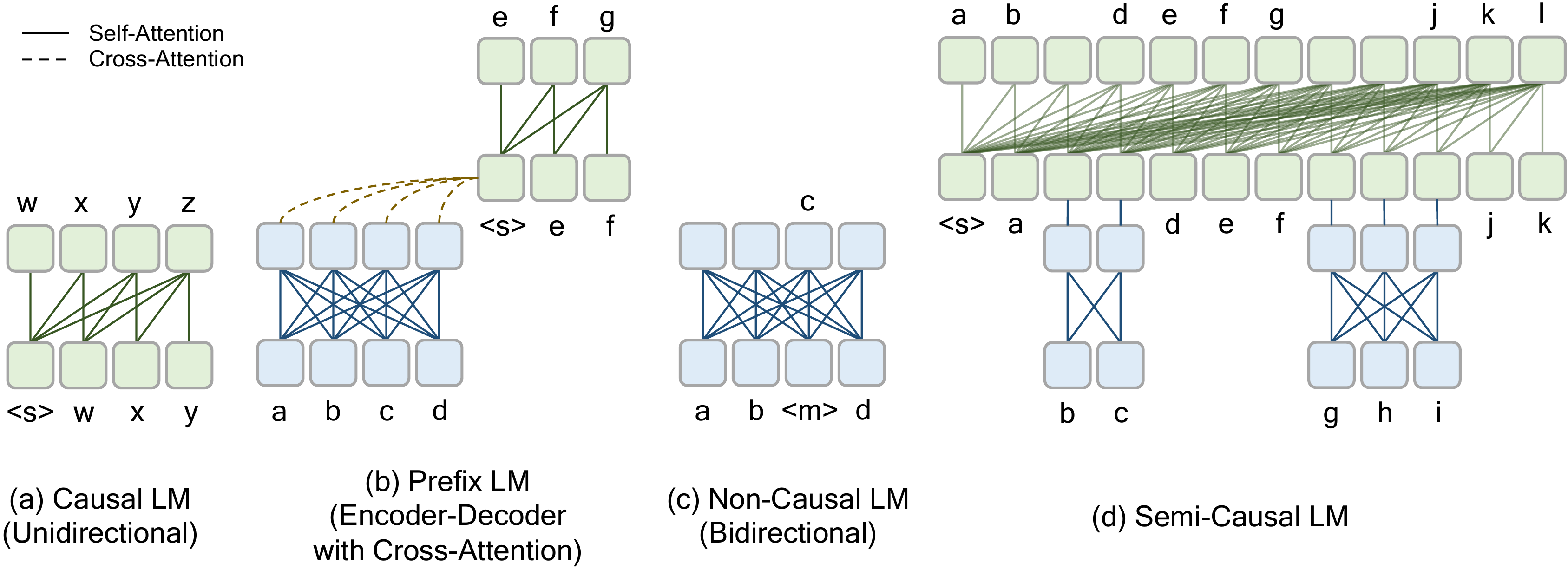}
\caption{Comparisons between different language model (LM) variants: (a) causal LM with unidirectional decoder~\citep{gpt3}; (b) prefix LM with encoder-decoder architecture~\citep{t5}; (c) non-causal LM with bidirectional encoder~\citep{bert}; (d) semi-causal LM proposed in this work.}
\label{fig:compare:lm}
\end{figure*}

As shown in Figure~\ref{fig:compare:lm}, we summarize the model architectures of three language model variants and the proposed semi-causal language model.
First, causal language model (such as GPT; \citealt{gpt3}) is a left-to-right Transformer decoder.
Second, prefix language model uses the encoder-decoder architecture with cross-attention connections to complete the sequence.
Third, non-causal language model is a bidirectional encoder, which is usually pretrained by masked language modeling~\citep{bert}.
Forth, the proposed semi-causal language model has a unidirectional Transformer decoder, and multiple bidirectional encoders that dock with the decoder.
In other words, our model processes the whole session from left to right, while having some spans pre-encoded by non-causal encoders.

\paragraph{Backbone Network}
We use Transformer~\citep{transformer} to build the models.
Given an input sequence, we first pack their vector representations together. Then we feed the vectors into a multi-layer Transformer, which encodes the input to contextualized representations.
In each Transformer block, there is a multi-head self-attention layer and a feed-forward network layer that are used to aggregate the hidden states of the previous layer.
Moreover, attention masks are used to control the context access.
We use a triangular matrix as the attention mask for the universal task layer, so that it processes the input from left to right.
For the bidirectional encoder, we allow all the tokens to access each other.
After obtaining the output vectors of the universal task layer, we use a softmax classifier to predict over the vocabulary.
The weight matrix is shared with the input token embeddings.

\paragraph{Connector}
As shown in Figure~\ref{fig:overview}, there is a connector layer between the universal task layer and various bidirectional encoders.
The connectors project vector representations of bidirectional encoders before feeding them into the general-purpose interface.
Moreover, the connectors are used to match the output dimensions of foundation models with the universal task layer.
We empirically find that both linear projection and feed-forward network work well in our experiments.

\subsection{Proposed Objective: Semi-Causal Language Modeling}
\label{sec:sclm}

In order to pretrain \ours{}, we introduce the semi-causal language modeling objective.
As shown in Figure~\ref{fig:overview}, our pretraining task autoregressively generates the tokens of a sequence, while some spans are represented by bidirectional encoders.

Given an input sequence $\bm{x}=x_1, x_2, ..., x_n$, we assume there are $k$ non-causal spans denoted as $\{\bm{x}_{s_1}^{e_1}, ..., \bm{x}_{s_k}^{e_k}\}$, where $\bm{x}_{s_i}^{e_i} = x_{s_i}, ..., x_{e_i - 1}$.
For each non-causal span $\bm{x}_{s_i}^{e_i}$, we use a bidirectional encoder to obtain its vector representations $\bm{h}( \bm{x}_{s_i}^{e_i} )$.
The choose of bidirectional encoders is dependent on the modality of the non-causal span.

Then the semi-causal language modeling objective is formulated as:
\begin{align}
\max \sum_{i=0}^{k} \sum_{t=e_i}^{s_{(i+1)}} \log P({x}_t | \bm{x}_{<{t}}, \{ \bm{h}(\bm{x}_{s_j}^{e_j}) \}_{j<i} )
\end{align}
where $e_0 = 1$, $s_{(k+1)} = n$, and $\{ \bm{h}(\bm{x}_{s_j}^{e_j}) \}_{j<i} = \{ \bm{h}(\bm{x}_{s_1}^{e_1}), \cdots, \bm{h}(\bm{x}_{s_{(i-1)}}^{e_{(i-1)}}) \}$.
Notice that the next token of each non-causal span is generated at the last position of the span.
Typically the number of non-causal spans and their positions are randomly sampled. The spans do not have overlaps with each other.

By leveraging the proposed objective, we jointly pretrain the general-purpose interface and the underlying foundational models, and seamlessly connect them together.
We pretrain \ours{} for both the language-only (Section~\ref{sec:lang:exp}) and vision-language (Section~\ref{sec:vl:exp}) settings.

\subsection{Capabilities on Downstream Tasks}

\paragraph{In-Context Learning}
\ours{} can adapt to a new task by conditioning on natural language instructions or several input-output pairs (i.e., demonstrations), without updating any parameter.
We first describe the usage of $k$-shot learning.
For each demonstration input, we conduct bidirectional encoding.
Then we feed the encoded vectors and the label into the general-purpose interface.
By conditioning on the given demonstrations, \ours{} predicts the target output of unseen examples.
For zero-shot generalization, there is only the test input, typically with prompts used to describe the task.
We feed the example with the task instruction into bidirectional encoders. The target output is generated by the universal task layer.

\paragraph{Finetuning}
Finetuning is especially helpful when many annotated examples of the downstream task are available.
We unify various tasks to the open-ended generation format, i.e., targets are transformed to free texts.
During finetuning, \ours{} learns to generate the target output, conditioning on the bidirectionally encoded input.
Compared with causal language models, \ours{} inherits the excellent finetuning capability of bidirectional encoders.

\paragraph{In-Context Customization}
A typical usage is that we first finetune the model on a large amount of data, and then use in-context learning to customize the finetuned model.
So we can easily transfer the knowledge of labeled data to new tasks.
As we subsume the advantages of both causal and non-causal modeling, \ours{} unlocks the combinations of the capabilities, i.e., good finetuning performance of non-causal modeling, and in-context learning of causal modeling.

\paragraph{Multimodal Multi-Turn Interaction}
\ours{} supports multi-turn interactions between users and pretrained models.
For each turn, non-causal modules encode user inputs, which accepts multimodal contents by using the corresponding pretrained encoders.
The output responses are generated by the general-purpose interface.
By conditioning on the history conversations, \ours{} naturally works as a conversational interface.
Moreover, the conversation can include multiple modalities instead of plain texts.

\section{Experiments on Language-Only Tasks}
\label{sec:lang:exp}

We first conduct experiments on language-only datasets to demonstrate the versatility and effectiveness of \ours{}.
Here the non-causal encoder is a pretrained language foundation model that docks with the universal task layer.
The intriguing capabilities emerge through pretraining, which enables the general-purpose interface to transfer across tasks and scenarios.

\subsection{Evaluation Settings}
\label{sec:lang:eval}

\begin{table}[t]
\centering
\renewcommand{\arraystretch}{1.3}
\begin{tabular}{l l}
\toprule
\multicolumn{1}{l}{\textbf{Evaluation Setting}} & 
\multicolumn{1}{l}{\textbf{Capability}} \\
\midrule
Multitask Finetuning & Perform a wide range of tasks competitively. \\
Single-Task Finetuning & Tackle individual tasks with remarkable performance. \\
Instruction Tuning & Zero-shot generalization after finetuning with instructions. \\
Zero-/Few-Shot Learning & Adapt to a new task given zero/few labeled examples. \\
\bottomrule
\end{tabular}
\caption{Summary of evaluation settings for language-only \ours{}. Each setting highlights an essential capability of \ours{}.}
\label{tbl:lang:eval}
\end{table}

We elaborate on language-only evaluation settings in  Table~\ref{tbl:lang:eval}.
We demonstrate the capabilities of \ours{}, including multitask finetuning (Section~\ref{sec:lang:multitask}), single-task finetuning (Section~\ref{sec:lang:ft}), instruction tuning (Section~\ref{sec:lang:instruct}), and in-context learning (Section~\ref{sec:lang:fewshot}).
The capabilities are task-agnostic and broadly applicable to understanding, generation, and interaction, which facilitates skill adaptation and communication with users.
Moreover, the evaluation settings of multitask finetuning and instruction tuning are seamlessly built upon the capability combination of finetuning and in-context learning.
In addition, because the tasks are unified in the free-text format, we can handle diverse downstream tasks using the same interface.

\begin{figure*}[t]
\centering
\includegraphics[width=\columnwidth]{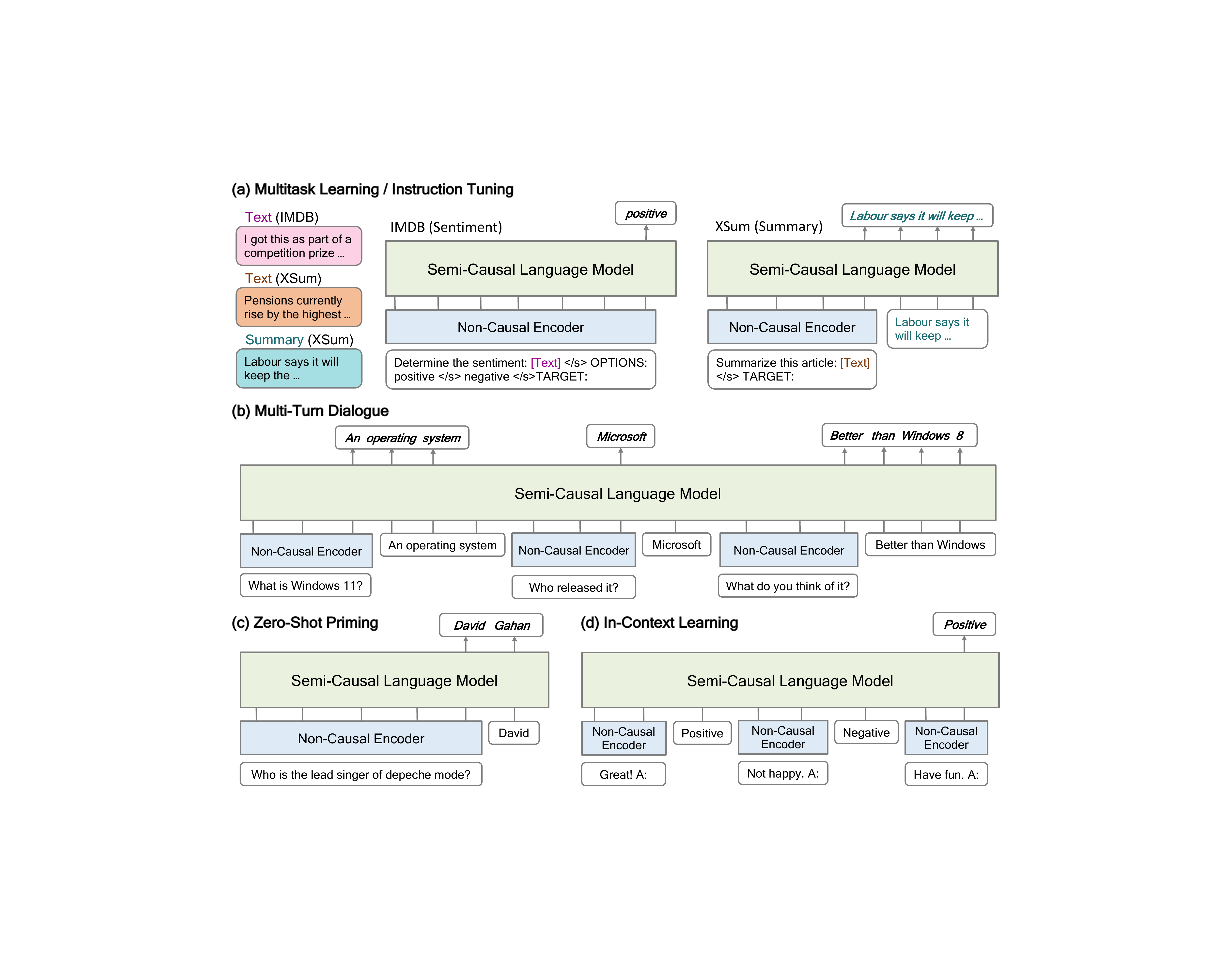}
\caption{\ours{} can be applied in different language-only scenarios: (a) multitask finetuning and instruction tuning, i.e., perform various tasks simultaneously in an open-ended manner.
(b) multi-turn dialogue, i.e., generate multi-turn responses according to the encoded input of users.
(c) zero-shot priming, e.g., natural question answering.
(d) few-shot learning, e.g., sentiment analysis.}
\label{fig:lang:eval}
\end{figure*}

Figure~\ref{fig:lang:eval} illustrates how to apply our model to different scenarios.
Generally, the input examples and instructions are fed to the non-causal language encoder, and the target outputs are produced from the universal task layer.
Moreover, the predictions are generated in a generative manner, which is open-ended.

\subsection{Pretraining Setup}
\label{sec:lang:setup}

We use sinusoidal position embeddings~\citep{transformer} for the language model.
The number of layers is $L=24$, each layer consists of $A=32$ attention heads and the hidden dimension is $H=2048$.
The number of parameters is about 1.3B.
For the non-causal part, we use encoder-only Transformers, where $A=16$, $H=1024$, $L=24$.
We utilize the learnable position embedding and relative position bias~\citep{t5} for the non-causal model.
The number of parameters is about 366M.
We use DeepNorm~\citep{deepnet} for Transformers.
The connector module is a linear projection layer in our implementation.

The maximum input lengths for non-causal and semi-causal models are 512 and 2048, respectively.
We randomly sample random spans whose lengths are between 64 and 128, and feed them to the non-causal part.
The total length of non-causal spans is 25\% of the original sequence length.
The spans do not cross document boundaries.
We pretrain the semi-causal language model from scratch.
The non-causal module is initialized from a pretrained bidirectional encoder, using the replaced token detection task~\citep{electra}.
During pretraining, we freeze all parameters of the non-causal encoder except the last two layers.
We pretrain \ours{} for 300k steps with a batch size of 1024 and use Adam~\citep{adam} for optimization.
We disable dropout of the semi-causal model and set the dropout rate of the non-causal model to 0.1.
We use a learning rate of 6e-4 with warm-up.
Please refer to Appendix~\ref{app:hyperparam:lang:pt} for more pretraining details.

We pretrain the model on Pile~\citep{pile}, which is a massive English text dataset constructed from diverse data sources and targeted at training large-scale language models.
We exclude data splits of GitHub, arXiv, and PubMed Central.
Please refer to Appendix~\ref{app:corpora:data:lang:pt} for detailed descriptions about Pile.
The pretraining data is tokenized by SentencePiece~\citep{sentencepiece}.
We construct the input in the ``full-sentence'' format~\citep{roberta}, i.e., each input sequence is packed with full sentences sampled contiguously from one or more documents.
We additionally introduce three special tokens for input construction: \texttt{<s>} indicates the start of a sequence, \texttt{</s>} indicates the end of a paragraph and \texttt{</d>} indicates the end of a document.

\subsection{Multitask Finetuning}
\label{sec:lang:multitask}

We first evaluate \ours{} under the multitask finetuning setting.
To be specific, we unify a wide range of tasks in an open-ended generation manner, so that they can be processed by the universal task layer without any task-specific architecture.
Figure~\ref{fig:lang:eval}(a) shows an example of how \ours{} handles multitask finetuning.
During finetuning, we randomly sample training examples and feed the inputs into the bidirectional language encoder. The finetuning objective is to maximize the likelihood of the correct labels generated from the interface.

We conduct experiments on a mixture of 34 NLP datasets (refer to Appendix~\ref{app:corpora:data:lang:ft} for more details) grouped into ten task clusters, including both language understanding tasks and generation tasks:
\begin{itemize}[leftmargin=*]
\item \textbf{Natural Language Inference}: ANLI (R1-R3), CB, MNLI, QNLI, RTE, SNLI, WNLI
\item \textbf{Sentiment Classification}: IMDB, SST-2, Sentiment140, Yelp
\item \textbf{Paraphrase Detection}: QQP, MRPC, Paws Wiki
\item \textbf{Coreference Resolution}: DPR, Winogrande, WSC
\item \textbf{Commonsense Reasoning}: HellaSwag, PiQA, COPA
\item \textbf{Reading Comprehension}: DROP, SQuADv1, SQuADv2, OBQA, BoolQ
\item \textbf{Miscellaneous}: CoLA, WiC, TREC
\item \textbf{Closed-Book QA}: ARC-easy, NQ
\item \textbf{Struct to Text}: CommonGen, E2ENLG
\item \textbf{Summarization}: AESLC, SamSum, XSum
\end{itemize}

\subsubsection{Evaluation Setup}
\label{sec:lang:multitask:detail}

\ours{} is finetuned on a mixture of all the mentioned datasets.
We limit the maximum number of training examples in each dataset to 30k.
We follow the prompts used in~\citep{flan}.
If the dataset is a multi-choice task, all possible options are provided in the template.
For instance, the input format of an example from a sentiment classification dataset is ``\texttt{<s>} \textit{Would the following phrase be considered positive or negative?} \texttt{</s>} [text] \texttt{</s>} \textit{OPTIONS:} \texttt{</s>} \textit{Positive} \texttt{</s>} \textit{Negative} \texttt{</s>} \textit{TARGET:}''.
The model determines the sentiment by generating \textit{Positive} or \textit{Negative}.

We finetune \ours{} for 20k steps with a batch size of 256.
The total length of input and answer tokens is restricted to 2048.
Following~\citep{t5}, we pack multiple training examples into one sequence to make computation batch-friendly.
The learning rate is set to 1e-4.
For more details, please refer to Appendix~\ref{app:hyperparam:lang:multi}.

For multi-choice tasks, we report the exact match score without decoding constraints.
For SQuAD, DROP, and closed-book QA datasets, we report the F1 score with greedy decoding.
When evaluating on the struct2text and summarization clusters, we use beam search~\citep{beam} with a beam size of 4 and a length penalty of $\alpha=0.6$.
We report ROUGE scores for the above two clusters.

\subsubsection{Results}

\begin{table}[t]
\centering
\begin{tabular}{c l c c}
\toprule
& {\textbf{Task Cluster}} & \textbf{GPT} & \textbf{\ours{}}  \\ \otoprule 
\multirow{7}*{\textbf{NLU}} 
& Natural Language Inference &  65.0 & \textbf{79.1}  \\
& Sentiment &  92.9 & \textbf{94.6} \\
& Paraphrase &  83.9 & \textbf{89.6}  \\
& Coreference &  67.1 & \textbf{84.3} \\
& Commonsense Reasoning &  63.3 & \textbf{84.2} \\
& Reading Comprehension &  64.5 & \textbf{73.1} \\
& Miscellaneous &  80.3 & \textbf{84.3} \\
\midrule
\multirow{3}*{\textbf{NLG}} 
& Closed-Book QA &  38.2 & \textbf{44.3} \\
& Struct to Text &  \textbf{44.2} & 44.1 \\
& Summarization &  29.8 & \textbf{31.0} \\
\bottomrule
\end{tabular}
\caption{Performance comparisons of multitask finetuning between \ours{} and GPT. We limit the number of training examples in each dataset to 30k during finetuning. For each task cluster, we present the average result over all sub-datasets within it. All results are reported on validation sets.}
\label{tbl:multitask_res}
\end{table}

\begin{figure*}[t]
\centering
\includegraphics[width=\columnwidth]{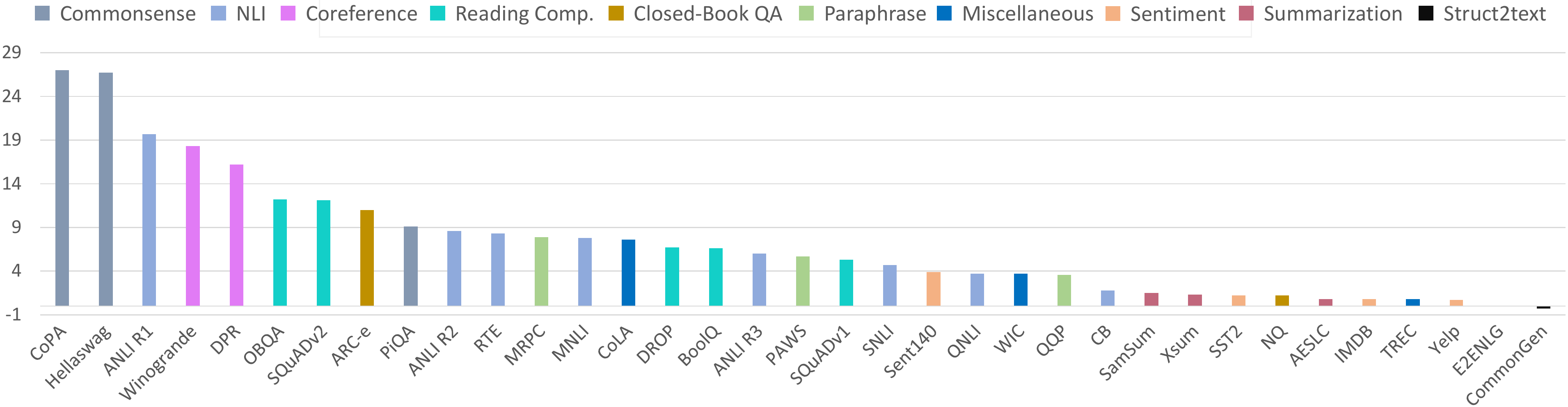}
\caption{Score difference of multitask finetuning results between \ours{} and GPT. We observe that \ours{} achieves consistent improvements over all tasks except the cluster of struct to text.}
\label{fig:lang:scorediff}
\end{figure*}

Table~\ref{tbl:multitask_res} compares the multitask finetuning results of \ours{} and GPT.
The GPT baseline follows the same configuration and training corpus for a fair comparison.
Each result represents the average score of all datasets of one task cluster.
The full results of all task clusters are reported in Appendix~\ref{app:results:lang:multitask}.
We also illustrate the score differences between \ours{} and GPT for all datasets in Figure~\ref{fig:lang:scorediff}.

We observe that \ours{} consistently surpasses GPT by a large margin on almost all the task clusters.
The results indicate that our method inherits the performant finetuning ability from the non-causal encoder.
Particularly, \ours{} performs much better than GPT on NLU tasks.
It partially confirms that non-causal modeling is conducive to finetuning~\citep{what:lm:objective,unifypara,role:bidirectionality}.
For more challenging tasks, such as natural language inference, and reading comprehension, the improvement of \ours{} is very prominent (14.1\% and 9.6\%).
Furthermore, we find that finetuning of GPT brings relatively small gains on commonsense reasoning tasks, whose results are comparable to zero-shot generalization.
By contrast, finetuning of \ours{} obtains decent gains over zero-shot numbers.
With regard to language generation, \ours{} consistently outperforms GPT except on struct-to-text datasets.
For closed-book question answering and text summarization, \ours{} achieves better performance than GPT too, benefiting from the non-causal modeling of input text.

\subsection{Single-Task Finetuning}
\label{sec:lang:ft}

We explore the finetuning capability of \ours{} under data-rich settings.
We design a new finetuning paradigm for \ours{}.
For each downstream task, we only update the parameters of the non-causal encoder while keeping the language model frozen.
We demonstrate that the proposed strategy achieves excellent performance, and preserves the general-purpose interface's capabilities of in-context learning and open-endedness.

\subsubsection{Finetuning Setup}

We conduct single-task finetuning on the natural language inference dataset MNLI~\citep{mnli}.
We use the template ``\texttt{<s>} \textit{Premise:}[*] \texttt{</s>} \textit{Hypothesis:}[*] \texttt{</s>} \textit{Label:}''.
The task is to determine whether a hypothesis is true, false or undetermined given a premise.
The corresponding labels are ``\textit{entailment}'', ``\textit{contradiction}'' and ``\textit{neutral}'', respectively.
During finetuning, we freeze the general-purpose interface and only update the non-causal encoder and the connector.
In contrast, all parameters are updated for the GPT baseline.
We finetune both \ours{} and GPT for three epochs with a learning rate of 5e-5 and a batch size of 32.

\subsubsection{Results}

\begin{table}[t]
\centering
\begin{tabular}{l c c}
\toprule
\multirow{2}*{\textbf{Model}} & \multicolumn{2}{c}{\textbf{MNLI} (acc)} \\ 
& -m & -mm  \\ \otoprule 
GPT & 87.7 & 87.6  \\
BERT~\citep{bert} & 86.6 & -    \\
RoBERTa~\citep{roberta} & 90.2 & 90.2    \\
ELECTRA~\citep{electra} & 90.9 & -    \\
\textbf{\ours{}} & \textbf{91.1} & \textbf{91.0}  \\
\bottomrule
\end{tabular}
\caption{Single-task finetuning results on matched (-m) and mismatched (-mm) validation sets of MNLI. Each score is the average of multiple runs with different random seeds.
}
\label{tbl:lang:ft_res}
\end{table}

Table~\ref{tbl:lang:ft_res} reports single-task finetuning accuracy.
MNLI-m and -mm represent the matched and the mismatched validation sets respectively.
Each score is the average of three runs with different random seeds.
Compared with GPT, \ours{} improves the accuracy of MNLI by 3.4 absolute points, despite updating much fewer parameters.
In addition to Section~\ref{sec:lang:multitask}, the results show that bidirectional encoders benefit finetuning performance~\citep{what:lm:objective,unifypara,role:bidirectionality}.
Furthermore, we also present three strong baselines derived from finetuning bidirectional language encoders, including BERT~\citep{bert}, RoBERTa~\citep{roberta} and ELECTRA~\citep{electra}.
All these three models are in large size.
Results show that \ours{} achieves comparable or better performance than the bidirectional encoders.

\subsection{Instruction-Tuned Zero-Shot Generalization}
\label{sec:lang:instruct}

We investigate instruction tuning for \ours{}, which finetunes the model on a variety of tasks with instructions.
After finetuning, we evaluate the performance of instruction following and zero-shot generalization for the models.
Because our goal is to investigate the zero-shot generalization on held-out tasks.
Therefore, when evaluating on a specific dataset, all datasets in the same category (i.e., task cluster) are not seen during the training stage.
For example, if we evaluate on the classification dataset SST-2, the entire cluster of sentiment analysis is excluded during instruction tuning.

\subsubsection{Instruction-Tuning Setup}

We follow the evaluation pipeline proposed in FLAN~\citep{flan}.
We conduct instruction tuning with \ours{} and GPT on the same dataset mixture described in Section~\ref{sec:lang:multitask} except for the summarization cluster.
For each dataset, we use ten different templates manually composed by FLAN~\citep{flan} and randomly apply one of them for every example.
As mentioned in~\citep{flan}, there are some templates that ``turned the task around'' to increase learning diversity, e.g., for sentiment classification, the model is prompted to generate a movie review based on the given sentiment label ``\textit{Positive}''.

Most finetuning configurations are the same as in Section~\ref{sec:lang:multitask:detail}.
We experiment on four task clusters, including natural language inference, sentiment classification, paraphrase detection, and reading comprehension.
Following the evaluation protocol of~\citep{flan}, the paraphrase cluster is dropped when evaluating on inference cluster and vice-versa.
We finetune \ours{} and GPT for 30k steps with a batch size of 512.
The learning rate is set to 1e-4.
The sequence length for each example is limited to 1024.
We also use the data packing strategy as in Section~\ref{sec:lang:multitask} to improve efficiency.
The detailed hyper-parameters is provided in  Appendix~\ref{app:hyperparam:lang:multi}.

\subsubsection{Results}

\begin{table}[t]
\centering
\begin{tabular}{@{}l c c c c}
\toprule
 & \multicolumn{2}{c}{\textbf{Avg template}} & \multicolumn{2}{c}{\textbf{Best template}} \\
 \cmidrule(r){2-3} \cmidrule(l){4-5}
 & GPT & \ours{} & GPT & \ours{} \\
 \midrule
 \multicolumn{5}{l}{\textit{Natural Language Inference}} \\
ANLI R1 & 31.6$_{0.6}$ & \textbf{36.2}$_{2.5}$ & 32.5 & \textbf{40.5} \\
ANLI R2 & 33.4$_{0.5}$ & \textbf{36.3}$_{1.2}$ & 34.0 & \textbf{38.2} \\
ANLI R3 & 35.9$_{1.3}$ & \textbf{38.9}$_{0.9}$ & 37.8 & \textbf{39.8} \\
CB & 60.3$_{4.3}$ & \textbf{75.0}$_{7.9}$ & 66.1 & \textbf{83.9} \\
MNLI-m & 45.8$_{2.6}$ & \textbf{51.0}$_{1.7}$ & 48.5 & \textbf{52.3} \\
QNLI & 59.3$_{0.7}$ & \textbf{66.1}$_{1.3}$ & 60.6 & \textbf{68.0} \\
RTE & 61.0$_{2.0}$ & \textbf{70.2}$_{3.3}$ & 64.3 & \textbf{75.5} \\
SNLI & 41.6$_{4.8}$ & \textbf{52.1}$_{4.3}$ & 49.8 & \textbf{58.1} \\
WNLI & 53.2$_{2.5}$ & \textbf{65.1}$_{3.9}$ & 56.3 & \textbf{71.8} \\
\textbf{Average} & 46.9 & \textbf{54.5} & 50.0 & \textbf{58.7} \\

 \midrule
 \multicolumn{5}{l}{\textit{{Sentiment}}} \\
IMDB & 84.6$_{2.6}$ & \textbf{85.8}$_{2.9}$ & 87.2 & \textbf{89.6} \\
SST-2 & 77.8$_{6.1}$ & \textbf{81.4}$_{6.4}$ & 83.9 & \textbf{89.9} \\
Sent140 & 85.4$_{1.1}$ & \textbf{86.4}$_{1.7}$ & 87.2 & \textbf{88.3} \\
Yelp & 84.1$_{10.8}$ & \textbf{91.0}$_{1.7}$ & \textbf{93.2} & 92.9 \\
\textbf{Average} & 83.0 & \textbf{86.2} & 87.9 & \textbf{90.2} \\

 \midrule
 \multicolumn{5}{l}{\textit{{Paraphrase}}} \\
QQP & \textbf{60.7}$_{0.7}$ & 59.7$_{2.1}$ & 61.6 & \textbf{62.1} \\
MRPC & 62.6$_{1.6}$ & \textbf{68.4}$_{0.5}$ & 65.2 & \textbf{69.1} \\
\textbf{Average} & 61.7 & \textbf{64.1} & 63.4 & \textbf{65.6} \\

 \midrule
 \multicolumn{5}{l}{\textit{{Reading Comprehension}}} \\
DROP & \textbf{18.1}$_{0.4}$ & 13.7$_{0.5}$ & \textbf{18.7} & 14.5 \\
SQuADv1 & 51.6$_{3.0}$ & \textbf{60.4}$_{1.5}$ & 55.6 & \textbf{62.7} \\
SQuADv2 & 24.9$_{1.3}$ & \textbf{28.7}$_{1.8}$ & 27.1 & \textbf{30.2} \\
OBQA & 28.4$_{1.3}$ & \textbf{36.2}$_{1.4}$ & 30.0 & \textbf{38.8} \\
BoolQ & 51.7$_{3.8}$ & \textbf{53.5}$_{2.6}$ & \textbf{57.8} & 56.7 \\
\textbf{Average} & 34.9 & \textbf{38.5} & 37.8 & \textbf{40.6} \\

\bottomrule
\end{tabular}
\caption{Full results of instruction tuning. We report the accuracy for all datasets except using F1 score for DROP, SQuADv1, and SQuADv2. The average score of each dataset is computed across five different templates.}
\label{tbl:flan_res}
\end{table}

Table~\ref{tbl:flan_res} reports the full results of instruction tuning on four task clusters.
For each dataset, we use five different templates for evaluation, and present both the average and the best score.
We observe that \ours{} achieves large improvements over the GPT baseline, which indicates the effectiveness of semi-causal language modeling.
Considering the natural language inference cluster, GPT fails to obtain reasonable zero-shot results on difficult datasets (such as ANLI and WNLI), while \ours{} consistently performs well on various datasets.
We notice similar trends on the other task clusters, i.e., sentiment, paraphrase, and reading comprehension.
In addition to the average results, \ours{} outperforms the GPT baseline in terms of the best performance.

The setting of instruction tuning requires the capabilities of both finetuning and zero-shot generalization.
Experimental results indicate that our method combines the best of causal and non-causal language models.
\ours{} not only achieves favorable finetuning performance because of bidirectional encoders, but also retains the causal language model's intriguing capability of zero-shot generalization.

\subsection{In-Context Learning}
\label{sec:lang:fewshot}

We compare the performance of in-context learning~\citep{gpt3} between \ours{} and GPT.
Conditioned on the task instruction and several input-label pairs, language models are repurposed towards the desired downstream task, following the input pattern while without updating parameters.
As illustrated in Figure~\ref{fig:lang:eval}(d), the demonstrations consist of two parts, the example input is passed through the non-causal encoder and the label token uses original embeddings.
Then the target label of the test input is generated by the universal task layer.

\subsubsection{Evaluation Setup}

We conduct experiments under zero-shot, one-shot, and four-shot settings.
We follow the evaluation protocol of GPT-3~\citep{gpt3}.
We evaluate each test example by randomly sampling examples from the training set as demonstrations.
The Winograd only has the test set, so we sample demonstrations directly from it.
Under few-shot settings, all examples are delimited by the separator token \texttt{</s>}.

We evaluate \ours{} and the GPT baseline on nine tasks, including cloze and completion tasks (i.e, StoryCloze, HellaSwag), Winograd-style tasks (i.e, Winograd, Winogrande), commonsense reasoning (i.e, ARC-easy, ARC-challenge, PIQA), and two datasets BoolQ and Copa from the SuperGLUE benchmark~\citep{superglue}.
The detailed descriptions of these datasets are provided in Appendix~\ref{app:corpora:data:lang:fewshot}.

\subsubsection{Results}

\begin{table}[t]
\centering
\begin{tabular}{l c c c c c c}
\toprule
\multirow{2}*{\textbf{Task}}
 & \multicolumn{2}{c}{\textbf{$k$=0}} & \multicolumn{2}{c}{\textbf{$k$=1}} & \multicolumn{2}{c}{\textbf{$k$=4}} \\
 \cmidrule(r){2-3} \cmidrule(l){4-5} \cmidrule(l){6-7}
 & {GPT} & \ours{} & {GPT} & \ours{} & {GPT} & \ours{} \\
 \midrule
StoryCloze & 72.4 & \textbf{73.1} & 72.5 & \textbf{74.2} & 72.5 & \textbf{73.6} \\
HellaSwag & 52.9 & \textbf{53.5} & 51.8 & \textbf{52.7} & 51.8 & \textbf{52.7} \\[0.2cm]

Winograd & 71.9 & \textbf{75.8} & 73.0 & \textbf{75.8} & 71.9 & \textbf{76.8} \\
Winogrande & \textbf{57.2} & 56.1 & 55.2 & \textbf{56.8} & \textbf{56.4} & \textbf{56.4} \\[0.2cm]

ARC-e & 50.6 & \textbf{52.6} & \textbf{53.1} & 51.1 & 54.3 & \textbf{56.1} \\
ARC-c & 28.8 & \textbf{31.2} & \textbf{28.5} & \textbf{28.5} & \textbf{29.5} & \textbf{29.5} \\
PIQA & \textbf{73.1} & 72.3 & \textbf{73.6} & 72.2 & \textbf{73.1} & 71.9 \\[0.2cm]

BoolQ & 62.1 & \textbf{62.2} & 57.6 & \textbf{57.9} & \textbf{61.5} & 61.3 \\
Copa & \textbf{70.0} & 67.0 & \textbf{69.0} & \textbf{69.0} & \textbf{71.0} & 70.0 \\
\midrule
\textbf{Average} & 59.9 & \textbf{60.4} & 59.4 & \textbf{59.8} & 60.2 & \textbf{60.9} \\
\bottomrule
\end{tabular}
\caption{Performance comparisons of in-context learning between \ours{} and GPT. $k$ represents the number of shots.}
\label{tbl:lang:few_shot}
\end{table}

Table~\ref{tbl:lang:few_shot} reports accuracy results of in-context learning.
Compared with GPT, \ours{} achieves better or comparable results.
For Winograd and completion tasks (i.e, StoryCloze, and HellaSwag), the performance of \ours{} has consistent improvements over GPT.
Considering the average result over these datasets, \ours{} is better in both zero-shot ($k=0$) and few-shot ($k=1,4$) settings.
The findings indicate that \ours{} inherits the excellent in-context learning ability, and the contextualized representations of non-causal encoders tend to help the model to generalize better.

\section{Experiments on Vision-Language Tasks}
\label{sec:vl:exp}

We conduct experiments under the vision-language setting.
The underlying non-causal encoder is a pretrained vision-language foundation model, which docks with the general-purpose interface.
The pretraining task is similar to the language-only setting, despite the use of image-text pairs.
Specifically, given an image-text pair, the image tokens are prepended to the text tokens.
As shown in Figure~\ref{fig:overview}, the non-causal encoder produces bidirectional fused representations of the image and a text prefix of random length.
The causal decoder is pretrained to autoregressively predict the remaining tokens conditioning on the bidirectional fused representations.
Text-only data is also leveraged and follows the same preparation protocol.
We jointly pretrain on both image-text data and text-only data during the vision-language \ours{} pretraining.

\subsection{Evaluation Settings}
\label{sec:vl:eval}

\begin{table}[t]
\centering
\resizebox{\textwidth}{!}{
\begin{tabular}{@{}lccccc@{}}
\toprule
\textbf{Dataset} & \textbf{Task description} & \textbf{Metric} & \textbf{Zero-shot} & \textbf{In-context} & \textbf{Finetuning} \\ \midrule
VQAv2         & Visual question answering    & VQA acc.    & \ding{51} & \ding{51} & \ding{51} \\
OK-VQA        & Knowledge-based VQA      & VQA acc.    & \ding{51} & \ding{51} & \ding{51} \\
VQA Karpathy  & Visual question answering    & VQA acc.    &           &           & \ding{51} \\
COCO Caption  & Image captioning      & CIDEr, etc. & \ding{51} &           & \ding{51} \\
Flickr30k Caption     & Image captioning      & CIDEr, etc. & \ding{51} &           &           \\
NoCaps        & Image captioning      & CIDEr, etc. & \ding{51} &           &           \\
NLVR$^2$      & Visual reasoning       & acc.        &           &           & \ding{51} \\
E-SNLI-VE label      & Visual reasoning       & acc.        &           &           & \ding{51} \\
E-SNLI-VE explanation  & Explanation generation & CIDEr, etc. &           &           & \ding{51} \\ \bottomrule
\end{tabular}
}
\caption{Evaluation summary of the vision-language datasets. 
We evaluate the capabilities of zero-shot, in-context learning, and finetuning.
}
\label{tab:vl:demo-capability}
\end{table}

Table~\ref{tab:vl:demo-capability} summarizes what capabilities we would like to evaluate and the corresponding vision-language datasets.
We conduct experiments on zero-shot generalization in~\cref{sec:vl:zeroshot}, in-context learning in~\cref{sec:vl:incontext}, and finetuning in~\cref{sec:vl:finetuning}.
The tasks can be grouped into several categories, i.e., visual question answering, visual reasoning, image captioning, and explanation generation.
The evaluation across nine datasets covers both understanding and generation.

\begin{figure*}[t]
\centering
\includegraphics[width=\textwidth]{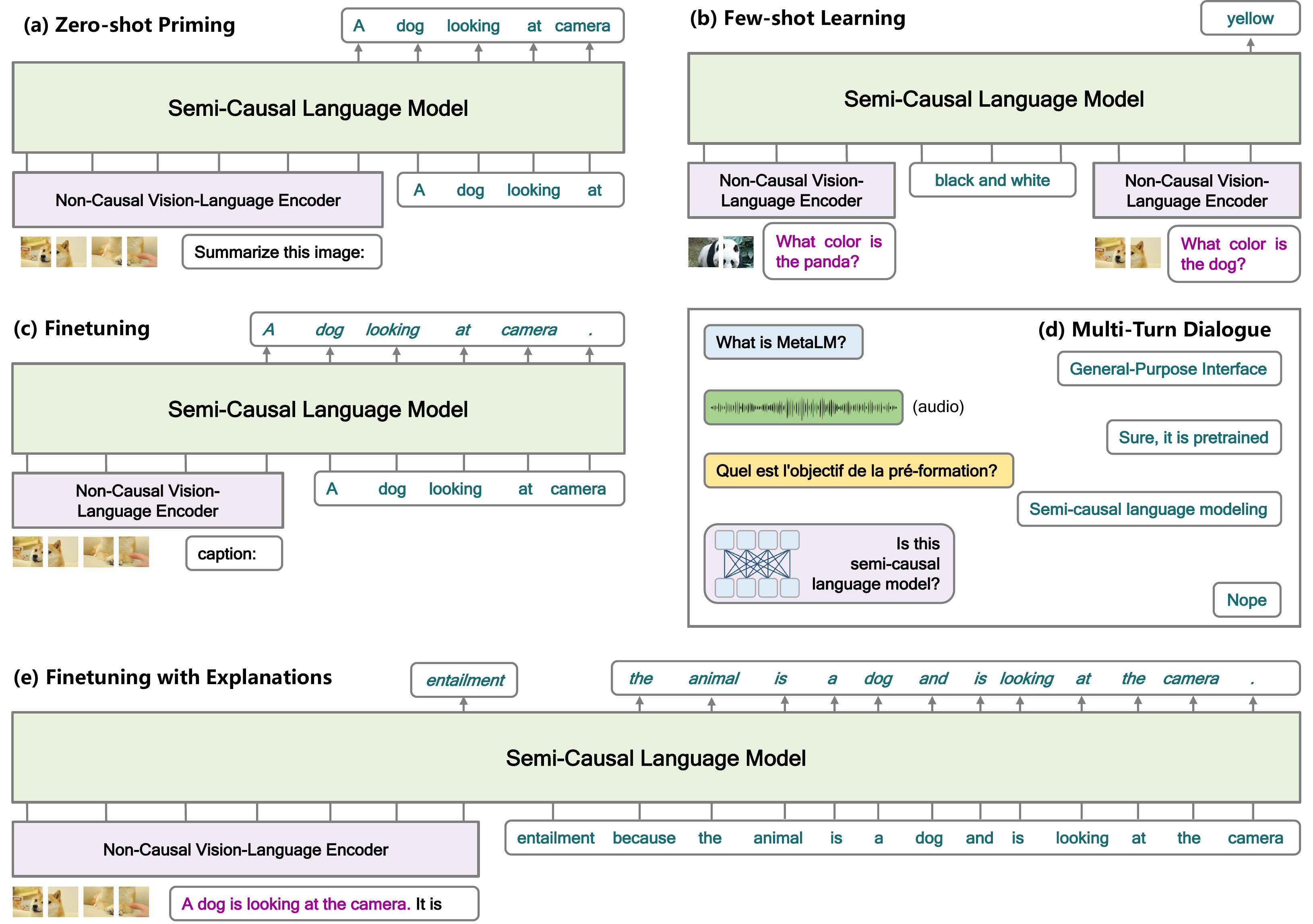}
\caption{The \ours{}'s capabilities include: (a) zero-shot priming, e.g., zero-shot image captioning with language prompts. (b) few-shot learning, e.g., visual question answering with in-context learning. (c) finetuning on different downstream tasks, e.g., image captioning, visual reasoning, etc. (d) multi-turn conversational interactions. (e) finetuning with explanations, i.e., using natural language explanations to guide the task learning.
}
\label{fig:vl:eval:protocol}
\end{figure*}

Figure~\ref{fig:vl:eval:protocol} illustrates how we evaluate \ours{} in different settings.
The input image and prompts are fed to a vision-language encoder, while the target output is generated by the language model.
All the tasks are formulated in an open-ended generative manner.

\subsection{Pretraining Setup}
\label{sec:vl:setup}

We use a 12-layer non-causal vision-language encoder and a 24-layer language model.
The universal task layer follows the same network architectures and configurations of GPT-2~\citep{gpt2}.
The hidden size is 1024, and there are 16 attention heads.
We employ sinusoidal position embeddings~\citep{transformer}.
The number of parameters is 353M.
For the non-causal encoder, we use a vision-language model pretrained as in VLMo~\citep{vlmo}.
The number of parameters is 192M.
We use 224x224 resolution during pretraining for images.
The connector is a three-layer feed-forward network.
More details about hyper-parameters can be found in Appendix~\ref{app:hyperparam:pt:vl}.

We pretrain \ours{} for 350k steps with 256 batch size. We use AdamW optimizer with $\beta_1=0.9$ and $\beta_2=0.98$. The learning rate is 1e-4 and weight decay is 0.01. We use linear decay and apply warm-up at the first 2,500 steps.
The dropout rate is set to 0.1.

We pretrain \ours{} using image-text pairs and text documents.
For image-text pairs, our pretraining data consists of Conceptual Captions~\citep{sharma2018conceptual}, Visual Genome~\citep{krishna2017vg}, COCO Caption~\citep{chen2015cococaption}, and SBU Caption~\citep{ordonez2011sbu} datasets.
Together, there are about 4M images and 10M image-text pairs.
For text documents, following \citep{roberta} and \citep{gpt2}, we use the OpenWebText~\citep{Aaron2019OpenWebText} corpus, which is an open-source recreation of the Reddit web text, as the pretraining data.

\subsection{Zero-Shot Generalization}
\label{sec:vl:zeroshot}

We evaluate the zero-shot generalization capability of \ours{} under vision-language settings.
Specifically, we conduct experiments on two tasks, including image captioning, and visual question answering.
For image captioning, only an input image is given, and the goal is to generate its description.
For visual question answering, a question is asked for the given image, and the model needs to predict the correct answers.

\subsubsection{Evaluation Setup}
\label{sec:Zero-shotEvaluationSetup}

We apply greedy decoding during inference.
The input images are resized to 224x224.
We describe the datasets and specific setups of two tasks as follows:

\paragraph{Image Captioning}
We evaluate zero-shot caption generation on MS COCO Caption~\citep{chen2015cococaption}, NoCaps~\citep{agrawal2019nocaps}, and Flickr30k~\citep{young2014flickr30k}.
We evaluate on the test set of COCO \textit{Karpathy split}~\citep{karpathy2017deep}, which re-partitions the train2014 and val2014 images~\citep{lin2014coco} into 113,287, 5,000, and 5,000 for train, validation, and test.
For NoCaps and Flickr30k, following \citep{jin2022fewVLM}, we evaluate on their validation set and test set, respectively.
We use BLEU~\citep{papineni2002bleu}, CIDEr~\citep{vedantam2015cider}, METEOR~\citep{banerjee2005meteor}, and SPICE~\citep{anderson2016spice} as caption generation metrics.
We utilize COCOEvalCap\footnote{\url{https://github.com/tylin/coco-caption}} to compute scores.
We prompt \ours{} with ``\textit{Summarize this image:}'' for all zero-shot caption generation experiments.

\paragraph{Visual Question Answering}
Following \citep{tsimpoukelli2021frozen}, we evaluate the zero-shot performance on VQAv2~\citep{vqav2} validation set and OK-VQA~\citep{marino2019okvqa} test set.
VQA score is calculated using normalization rules of the VQAv2 evaluation code.\footnote{\url{https://github.com/GT-Vision-Lab/VQA}}
Different from classification over a predefined set of candidate answers, \ours{} predicts answers in an open-ended generation manner.
We prompt \ours{} with the template``question: \textit{question text} answer:'' for all visual question answering experiments.

\subsubsection{Results}

\begin{table}[t]
\centering
\begin{tabular}{@{}lcccc@{}}
\toprule
 & \multicolumn{4}{c}{\textbf{COCO Caption Karpathy Test}} \\ \cmidrule(l){2-5} 
\multirow{-2}{*}{\textbf{Model}} & BLEU-4 & CIDEr & METEOR & SPICE \\ \midrule
{ZeroCap~\citep{tewel2021zero}} & {2.6} & {14.6} & { 11.5} & {5.5} \\
{VLKD$_\text{ViT-B/16}$~\citep{dai2021enabling}} & {16.7} & {58.3} & {19.7} & {13.4} \\
\textbf{\ours{}} & \textbf{24.5} & \textbf{82.2} & \textbf{22.5} & \textbf{15.7} \\
\bottomrule
\end{tabular}
\caption{Zero-shot generalization on COCO image captioning.
}
\label{tab:zs-caption:coco}
\end{table}

\begin{table}[t]
\centering
\begin{tabular}{@{}lcccc@{}}
\toprule
\multirow{2}{*}{\textbf{Model}} & \multicolumn{2}{c|}{\textbf{NoCaps}} & \multicolumn{2}{c}{\textbf{Flickr30k}} \\ \cmidrule(l){2-5} 
 & CIDEr & SPICE & CIDEr & SPICE \\ \midrule
VL-T5~\citep{cho2021unifying} & 4.4 & 5.3 & 2.6 & 2.0 \\
FewVLM~\citep{jin2022fewVLM} & 42.2 & 8.5 & 31.0 & 10.0 \\
\textbf{\ours{}} & \textbf{58.7} & \textbf{8.6} & \textbf{43.3}  & \textbf{11.7} \\ \bottomrule
\end{tabular}
\caption{Zero-shot image captioning results on NoCaps validation and Flickr30k test. All the results are from their \textit{base} size models, and the numbers are taken from \citep{jin2022fewVLM}.
}
\label{tab:zs-caption}
\end{table}

Table~\ref{tab:zs-caption:coco} and Table~\ref{tab:zs-caption} show the zero-shot captioning results on COCO Karpathy test split, NoCaps validation set, and Flickr30k test set.
\ours{} outperforms recent strong methods on three image captioning datasets.
To be specific, the compared model FewVLM~\citep{jin2022fewVLM} leverages different prompts for image captioning, and we report its best results.
By contrast, we use the same prompt ``\textit{Summarize this image:}'' for comparisons in all the experiments.
Our model robustly follows the instruction to produce readable captions in the zero-shot manner.

\begin{table}[t]
\centering
\begin{tabular}{@{}l cc@{}}
\toprule
{\textbf{Model}} & {\textbf{VQAv2}} & {\textbf{OK-VQA}} \\
\midrule
Frozen~\citep{tsimpoukelli2021frozen} & 29.5 & 5.9  \\
VLKD$_\text{ViT-B/16}$~\citep{dai2021enabling} & 38.6 & 10.5 \\
\textbf{\ours{}} & \textbf{41.1} & \textbf{11.4} \\
\bottomrule
\end{tabular}
\caption{Zero-shot generalization on visual question answering. All models predict in a generative manner without additional information, such as captions and object tags.}
\label{tab:vl:zero-shot-vqa}
\end{table}

Table~\ref{tab:vl:zero-shot-vqa} reports the results of zero-shot visual question answering on VQAv2 and OK-VQA.
On both datasets, \ours{} achieves better zero-shot results than Frozen~\citep{tsimpoukelli2021frozen} and VLKD~\citep{dai2021enabling}, even though Frozen has significantly more parameters.
In addition, the OK-VQA dataset is designed for visual question answering that is supposed to require external knowledge.
For example, the input image is a train, and the asked question is ``\textit{When is it invented?}''.
The reasonable performance on OK-VQA indicates that the language model of \ours{} tends to serve as a knowledge source.
Once object information is perceived by the vision encoder, the universal task layer generates the answer as language modeling.

The experimental results across five datasets show that \ours{} has the capabilities of zero-shot generalization and open-ended generation.
We can use prompts to re-purpose the pretrained vision-language model to image captioning and visual question answering.

\subsection{In-Context Learning}
\label{sec:vl:incontext}

We evaluate the capability of in-context learning~\citep{gpt3} on visual question answering.
We conduct $k$-shot learning, where $k$ demonstrations are used to guide the prediction of new examples without finetuning the parameters.

\subsubsection{Evaluation Setup}

Following~\citep{tsimpoukelli2021frozen}, we carry out few-shot experiments on the VQAv2~\citep{vqav2} validation set and OK-VQA~\citep{marino2019okvqa} test set.
We randomly sample up to four full examples from the training set for each test instance.
The predicted answers are evaluated against the ground-truth answers following the normalization rules from the VQAv2 evaluation code.
We use an image resolution of 224x224 during inference.

As shown in Figure~\ref{fig:vl:eval:protocol}(b), we put several examples before the test input and directly obtain the prediction from the universal task layer.
Specifically, a full example is denoted as $e=[i, q, a]$, where \textit{i}, \textit{q}, \textit{a} denote image, question, and answer, respectively.
Similarly, a test input \textit{t} is denoted as $t=[i, q]$.
For $k$-shot in-context learning, the whole input sequence is $e_1 , ... , e_k , t$.
Moreover, we use ``\textit{Question: [question text] Answer:}'' as the prompt to instruct \ours{}.
Then \ours{} uses greedy decoding to generate answers.

\subsubsection{Results}

\begin{table}[t]
\centering
\begin{tabular}{@{}l cc cc@{}}
\toprule
\multirow{2}{*}{\textbf{Model}} & \multicolumn{2}{c|}{\textbf{VQAv2}} & \multicolumn{2}{c}{\textbf{OK-VQA}} \\ \cmidrule(l){2-5} 
 &  $k$=1 & $k$=4 & $k$=1 & $k$=4 \\ \midrule
Frozen~\citep{tsimpoukelli2021frozen} & 35.7 & 38.2 & 9.7 & 12.6 \\
\textbf{\ours{}} & \textbf{42.4} & \textbf{45.3} & \textbf{13.2} & \textbf{16.0} \\
\bottomrule
\end{tabular}
\caption{In-context learning on visual question answering. All models predict in a generative manner without additional information, such as captions and object tags. $k$ is the number of in-context examples~\citep{gpt3} that the model can learn from.}
\label{tab:vl:few-shot-vqa}
\end{table}

Table~\ref{tab:vl:few-shot-vqa} reports the in-context learning results on the visual question answering datasets VQAv2 and OK-VQA.
The results show that adding in-context demonstrations improves the performance over zero-shot generalization as shown in Table~\ref{tab:vl:zero-shot-vqa}.
Besides, adding more examples brings larger improvements to both datasets.
Compared with Frozen~\citep{tsimpoukelli2021frozen}, \ours{} obtains better performance despite the use of relatively small model size.
We find that \ours{} can conduct in-context learning on visual question answering without modifying the underlying vision-language model.
Although the non-causal encoder only sees one example each time, the language model successfully adapts the model according to the $k$ demonstrations.
In addition, with the help of the universal task layer, we can augment the existing foundation models with the general capability of in-context learning.

\subsection{Finetuning on Downstream Tasks}
\label{sec:vl:finetuning}

We finetune the pretrained \ours{} on a wide range of vision-language tasks, including image captioning~\citep{karpathy2017deep}, visual question answering~\citep{vqav2,marino2019okvqa}, visual reasoning~\citep{nlvrv2}, and explainable visual reasoning~\citep{kayser2021vil}.
We compare the finetuned \ours{} with both the strong discriminative models and recent generative models.

\subsubsection{Finetuning Setup}

For all tasks, we use the resolution of 384x384 during finetuning.
We also apply RandAugment~\citep{cubuk2020randaugment} for image augmentation.
We keep the learning rate 1e-5 fixed for all datasets.
More detailed hyper-parameters can be found at Appendix~\ref{app:hyperparam:ft:vl}.
We describe the setups of various tasks as follows.

\paragraph{Visual Question Answering}
We evaluate on VQAv2~\citep{vqav2}, VQA Karpathy split~\citep{cho2021unifying}, and OK-VQA~\citep{marino2019okvqa}.
For VQAv2, models are finetuned on the training and validation sets. We report the VQA score on the \textit{test-dev} and \textit{test-std} sets.
For VQA Karpathy split, models are finetuned on the training and validation sets. We report the VQA score on the \textit{in-domain} and \textit{out-domain} test set.
We finetune \ours{} for 140k steps for both the above two datasets.
For OK-VQA, models are finetuned on the training set. We report the normalized VQA score on the test set. We finetune \ours{} with 10k steps.
We apply a ``Question: [\textit{question text}] Answer: [\textit{answer text}]'' prompt for generative finetuning.

\paragraph{Visual Reasoning}
We evaluate on the NLVR$^2$ dataset~\citep{nlvrv2}. The example in NLVR$^2$ consists of two images and one sentence, where the sentence describes the relations between the images. Following previous work~\citep{tan2019lxmert,li2020oscar}, we re-split the data into two individual image-text pairs and get their representations respectively. Then we leverage the concatenation of representations to generate the \textit{yes} or \textit{no} predictions.
We apply ``it is [\textit{label}]'' for generative finetuning.
We finetune \ours{} for 5 epochs.

\paragraph{Image Captioning}
We evaluate on the COCO caption dataset with \textit{Karpathy split}~\citep{karpathy2017deep}. Following~\citep{cho2021unifying}, we report BLEU-4, CIDEr, METEOR, and SPICE as the evaluation metrics. All reported results are from cross-entropy finetuning without reinforced CIDEr optimization~\citep{rennie2017self}. Object tags are not used during finetuning.
We apply a ``caption: [\textit{caption text}]'' prompt for generative finetuning and finetune \ours{} for 100k steps on the training split.

\paragraph{Explainable Visual Reasoning}
We evaluate on the E-SNLI-VE dataset~\citep{kayser2021vil}, which requires the models to predict the entailment labels between an image-text pair and simultaneously generate explanations for the prediction.
We finetune \ours{} for 7 epochs.
This task is naturally compatible with the language generation manner. We apply a ``it is [\textit{entailment label}] because [\textit{explanation}].'' prompt for generative finetuning.

\subsubsection{Results: Visual Question Answering and Visual Reasoning}

\begin{table}[t]
\centering
\begin{tabular}{@{}lccccc@{}}
\toprule
\multirow{2}{*}{\textbf{Model}} & \multicolumn{2}{c}{\textbf{VQAv2}} & \multicolumn{2}{c}{\textbf{VQA Karpathy-test}} & \textbf{NLVR$^2$} \\ \cmidrule(l){2-6} 
& test-dev & test-std & In-domain & Out-domain & test-P \\
\midrule
\multicolumn{6}{l}{\textit{Discriminative Prediction}} \\
ViLBERT~\citep{lu2019vilbert} & 70.6 & 70.9 & - & - & - \\
Oscar~\citep{li2020oscar} & 73.2 & 73.4 & - & - & 78.4 \\
UNITER~\citep{chen2020uniter} & 72.3 & 72.9 & 74.4 & 10.0 & 77.9 \\
\midrule
\multicolumn{6}{l}{\textit{Generative Prediction}} \\
VL-T5~\citep{cho2021unifying} & - & 70.3 & 71.4 & 13.1 & 73.6 \\
VL-BART~\citep{cho2021unifying} & - & 71.3 & 72.1 & 13.2 & 70.3 \\
VLKD$_\text{ViT-B/16}$~\citep{dai2021enabling} & 69.8 & - & 69.2 & 18.6 & - \\
\textbf{\ours{}} & \bf 74.4 & \bf 74.5 & \bf 77.9 & \bf 21.1 & \bf 80.9 \\
\bottomrule
\end{tabular}
\caption{Comparison of finetuning results on different vision-language tasks.
The discriminative manner predicts a distribution over a pre-defined set of labels, e.g., 3129 most common answers for VQAv2. In contrast, the open-ended generative manner handles all tasks with free-text generation.
Notice that all the reported results are from their \textit{base} size models.}
\label{tab:vl:ft:vqa-nlvr}
\end{table}

Table~\ref{tab:vl:ft:vqa-nlvr} reports the finetuning results on VQAv2, VQA Karpathy, and NLVR$^2$.
The finetuning performance is strong across the datasets.
More importantly, \ours{} not only outperforms previous models with generative prediction, but also achieves competitive or better results compared with discriminative vision-language models.
The property is favorable as the nature of some tasks is generative. For example, visual question answering needs open-ended predictions, rather than restricting the output space.
The advantages of open-endedness are shown on the out-domain set of the VQA Karpathy-test.
The top answers of the out-domain set are not in the most common 3,129 VQA answers.
As the discriminative models can only make predictions that appear in the training set, it is difficult to generalize to out-domain examples.
Among all the models, \ours{} achieves the best out-domain results.
In comparison, although previous generative models get better results on the out-domain set, they usually underperform on other datasets.
By contrast, \ours{} consistently achieves competitive results.

\begin{table}[t]
\centering
\begin{tabular}{@{}lc}
\toprule
\textbf{Model} & \textbf{OK-VQA} \\
\midrule
\multicolumn{2}{l}{\textit{Discriminative Prediction}} \\
ViLBERT~\citep{lu2019vilbert} & 35.2 \\
KRISP~\citep{marino2021krisp} & 38.9 \\
MAVEx~\citep{wu2022MAVEx} & 40.3 \\ \midrule
\multicolumn{2}{l}{\textit{Generative Prediction}} \\
VLKD$_\text{ViT-B/16}$~\citep{dai2021enabling} & 36.3 \\
\textbf{\ours{}} & \textbf{46.5} \\
\bottomrule
\end{tabular}
\caption{Finetuning results on the knowledge-intensive OK-VQA dataset. Different from the VQAv2,
this dataset requires not only understanding images and questions but also leveraging world knowledge. For example, for an image of a plane, the question is ``\textit{who invented this?}''.
All the reported results are taken from their \textit{base} size models.}
\label{tab:vl:ft:okvqa}
\end{table}

As shown in Table~\ref{tab:vl:ft:okvqa}, we report the finetuning results on OK-VQA~\citep{marino2019okvqa}.
Different from VQAv2, the dataset requires models to draw upon external knowledge to answer questions.
Previous methods~\citep{marino2021krisp,wu2022MAVEx} typically leverage a knowledge base to filter candidate answers.
In contrast, language models have acquired rich world knowledge during pretraining. \ours{} grants the flexibility of leveraging such knowledge from the causal language model.
As a result, \ours{} obtains significant improvements on this task without relying on additional knowledge bases.

\begin{table}[t]
\centering
\begin{tabular}{@{}lc@{}}
\toprule
\textbf{Model} & \textbf{Accuracy} \\
\midrule
\citep{park2018multimodal} & 69.2 \\
\citep{wu-mooney-2019-faithful} & 73.7 \\
\citep{marasovic-etal-2020-natural} & 72.0 \\
\citep{kayser2021vil} & 79.5 \\
\citep{nlx-gpt} & 73.9 \\
\textbf{\ours{}} & \textbf{79.9} \\
\quad w/o appending explanations after labels & 79.6 \\
\bottomrule
\end{tabular}
\caption{Comparison of finetuning results on E-SNLI-VE~\citep{kayser2021vil}.
Without explanation \ours{} still predicts the entailment label in an open-ended generative manner.
The compared results are taken from \citep{kayser2021vil} and \citep{nlx-gpt}.
}
\label{tab:vl:ft:e-snli-ve:nlu}
\end{table}

Table~\ref{tab:vl:ft:e-snli-ve:nlu} reports the finetuning results on E-SNLI-VE entailment label prediction.
\ours{} is trained to jointly generate the entailment label and explanation with the ``it is [\textit{entailment label}] because [\textit{explanation}]'' prompt.
\ours{} achieves the best accuracy compared with previous methods.
Moreover, an important advantage of the generative model is that \ours{} can leverage explanations to improve the performance of entailment label prediction.
It indicates that the explanation is of help to entailment classification.
The results demonstrate that \ours{} can be used to facilitate the interactions between users and foundation models.
In other words, we can use natural language to guide model finetuning via the general-purpose interface.

The competitive results across the above datasets demonstrate that the bidirectional modeling benefits finetuning in \ours{}.
So we can have good performance of finetuning and open-ended prediction at the same time.

\subsubsection{Results: Visually Grounded Language Generation}

\begin{table}[t]
\centering
\begin{tabular}{@{}lcccc@{}}
\toprule
\multirow{2}{*}{\textbf{Model}} & \multicolumn{4}{c}{\textbf{COCO Caption Karpathy Test}} \\ \cmidrule(l){2-5} 
 &  BLEU-4 & CIDER & METEOR & SPICE \\
\midrule
Oscar~\citep{li2020oscar} & 34.5 & 115.6 & 29.1 & 21.9 \\
Unified VLP~\citep{zhou2020unified} & 36.5 & 117.7 & 28.4 & 21.3 \\
VL-T5~\citep{cho2021unifying} & 34.6 & 116.1 & 28.8 & 21.9 \\
VL-BART~\citep{cho2021unifying} & 34.2 & 114.1 & 28.4 & 21.3 \\
\textbf{\ours{}} & \textbf{37.6} & \textbf{126.6} & \textbf{30.0} & \textbf{22.9} \\ \bottomrule
\end{tabular}
\caption{Finetuning results on the COCO caption Karparthy test split. All models are directly finetuned without using CIDEr optimization~\citep{rennie2017self} and object tags. The results of base-size models are taken from~\citep{cho2021unifying}.
}
\label{tab:vl:ft:coco-caption}
\end{table}

Table~\ref{tab:vl:ft:coco-caption} reports the finetuning results of caption generation on COCO Karpathy test split.
We directly compare with the results without CIDEr optimization~\citep{rennie2017self} for fair comparisons.
The results show that \ours{} obtains substantial improvements over other models.

\begin{table}[t]
\centering
\resizebox{\textwidth}{!}{
\begin{tabular}{@{}llllllll}
\toprule
\textbf{Model} & \multicolumn{1}{c}{\textbf{BLEU-1}} & \multicolumn{1}{c}{\textbf{BLEU-2}} & \multicolumn{1}{c}{\textbf{BLEU-3}} & \multicolumn{1}{c}{\textbf{BLEU-4}} & \multicolumn{1}{c}{\textbf{ROUGE-L}} & \multicolumn{1}{c}{\textbf{METEOR}} & \multicolumn{1}{c}{\textbf{CIDER}} \\ \midrule
\citep{park2018multimodal} & 29.4 & 18.0 & 11.3 & 7.3 & 28.6 & 14.7 & 72.5 \\
\citep{wu-mooney-2019-faithful} & 30.6 & 19.2 & 12.4 & 8.2 & 29.9 & 15.6 & 83.6 \\
\citep{marasovic-etal-2020-natural} & 29.9 & 19.8 & 13.6 & 9.6 & 27.3 & 18.8 & 81.7 \\
\citep{kayser2021vil} & 30.1 & 19.9 & 13.7 & 9.6 & 27.8 & \textbf{19.6} & 85.9 \\
\citep{nlx-gpt} & 37.0 & 25.3 & 17.9 & 12.9 & 34.2 & 18.8 & 117.4 \\
\textbf{\ours{}} & \textbf{40.6} & \textbf{26.7} & \textbf{18.7} & \textbf{13.5} & \textbf{37.6} & 19.4 & \textbf{119.3} \\ \bottomrule
\end{tabular}
}
\caption{Finetuning results of E-SNLI-VE explanation generation. \ours{} jointly generates entailment labels and explanations.
The compared results are taken from \citep{nlx-gpt}.
}
\label{tab:vl:ft:e-snli-ve:nlg}
\end{table}

Table~\ref{tab:vl:ft:e-snli-ve:nlg} shows the explanation generation results on E-SNLI-VE.
We jointly generate entailment labels and explanations.
\ours{} outperforms previous strong models on most metrics.
Together with the label accuracy results on the same dataset in Table~\ref{tab:vl:ft:e-snli-ve:nlu}, our model achieves good performance for both understanding and explanation generation.
In contrast, the method of \citep{nlx-gpt} obtains competitive performance for explanation generation, while getting inferior accuracy for entailment classification.

The results of visually grounded language generation show that our architecture is general enough to be applied to various sequence-to-sequence learning problems.
\ours{} can achieve good performance via finetuning for vision-language generation tasks.

\section{Related Work}
\label{sec:related}

\subsection{Language Model Pretraining}

Large-scale language model pretraining has achieved strong performance across various downstream tasks and aroused extensive research interest.
The difference between the models mainly lies in the pretraining objective and model architecture.
GPT~\citep{gpt1,gpt2,gpt3} pretrains causal language models with decoder-only Transformers, demonstrating intriguing properties of few-shot and in-context learning.
Recent efforts~\citep{gopher,glam,mt-nlg,chinchilla,lamda,palm} focus on scaling up in terms of data and model size.
In order to implement bidirectional encoding, \citet{bert} propose the masked language modeling objective.
\citet{electra} introduce the replaced token detection task to improve pretraining efficiency.
Furthermore, some efforts investigate frameworks that can handle both natural language understanding and generation tasks.
T5~\citep{t5} introduces an encoder-decoder framework that converts all tasks into a text-to-text format.
BART~\citep{bart} is a sequence-to-sequence model pretrained by reconstructing the original text from corrupted documents.
UniLM~\citep{unilm,unilmv2} presents to jointly optimize unidirectional, bidirectional and sequence-to-sequence language modeling objectives controlled by different self-attention masks.
\citet{what:lm:objective}, \citet{unifypara}, and \citet{role:bidirectionality} study the effects of different pretraining objectives and architectures on downstream generalization.
Specifically, causal language models are good at zero-shot or in-context learning, while non-causal models perform better for finetuning.
In our work, we combine the best of both worlds by introducing semi-causal language modeling. So we can obtain decent finetuning performance and benefit from the capability of in-context learning.
Moreover, the unification enables us to build a general-purpose interface to various foundation models.

\subsection{General-Purpose Modeling}

Some efforts investigate the general-purpose model that supports multiple tasks, transformations, and modalities in a shared module.
MT-DNN~\citep{mtdnn} trains on many tasks through multitask learning.
Specific to language-only general-purpose, UniLM~\citep{unilm} and T5~\citep{t5} unify understanding and generation ability in a single model.
Moreover, language models are finetuned to follow instructions~\citep{instructgpt,flan,tzero}, i.e., aligning language models with user intentions to implement the general-purpose capability.
There are some work that support not only multitask but also multimodality.
\citet{perceiverio} introduce Perceiver IO, a general architecture across multiple domains including language/visual understanding, multimodal and symbolic representations for games.
\citet{data2vec} propose a unified learning framework for different modalities but still use modality specific encoders.
\citet{tsimpoukelli2021frozen} demonstrate that the in-context learning ability of frozen language models can be transferred to a vision-language setting.
\citet{flamingo} also implement general-purpose understanding of image, video, and text by a large frozen language model.
\citet{gato} build a generalist agent that works as a multi-modal, multi-task, multi-embodiment generalist policy.

\section{Conclusion}
\label{sec:conclusion}

We present \ours{}, a general-purpose interface to foundation models across tasks and modalities.
\ours{} consists of a causal decoder as the universal task layer, and multiple pretrained non-causal encoders mounted to it.
We pretrain \ours{} with a new objective called semi-causal language modeling.
Experimental results show that \ours{} exhibits strong finetuning and in-context learning performance across a wide range of language-only and vision-language tasks.

In the future, we would like to scale up~\citep{deepnet,xmoe} the model size.
Moreover, we are interested in extending \ours{} to multilingual settings, and handling more modalities (including language, vision, audio, and multimodality) simultaneously.
Another strand of work is to extend the universal task layer to vision tasks, such as object detection, and semantic segmentation.
We will also investigate parameter-efficient finetuning with \ours{}.

\bibliographystyle{plainnat}
\bibliography{metalm}

\newpage

\appendix

\section{Hyperparameters of Language-Only Experiments}
\label{app:hyperparam:lang}
\subsection{Pretraining}
\label{app:hyperparam:lang:pt}
We provide the detailed pretraining hyperparameter settings of language-only \ours{}.
Model hyperparameters are shown in Table~\ref{tbl:hyperparam:lang:pt:model} and optimization hyperparamters are shown in Table~\ref{tbl:hyperparam:lang:pt:opt}.

\noindent\hspace{0.03\linewidth}
\begin{minipage}[c]{0.5\textwidth}
\centering
\begin{tabular}{lcc}
\toprule
\textbf{Hyperparameters} & \textbf{Non-causal} & \textbf{Semi-causal} \\ \midrule
Number of layers & 24 & 24 \\
Hidden size & 1024 & 2048 \\
FFN inner hidden size & 4096 & 8192 \\
Attention heads & 16 & 32 \\
Attention head size & 64 & 64 \\
Dropout & 0.1 & 0.0 \\
Attention Dropout & 0.1 & 0.0 \\
Initialization & DeepNorm & DeepNorm \\
Max length & 512 & 2048 \\
Position Embedding & Learnable & Sinusoidal \\
\bottomrule
\end{tabular}
\captionsetup{type=table}
\caption{Hyperparameters of non-causal and semi-causal models for language-only pretraining.}
\label{tbl:hyperparam:lang:pt:model}
\end{minipage}
\hfill
\begin{minipage}[c]{0.4\textwidth}
\centering
\begin{tabular}{lr}
\toprule
\textbf{Hyperparameters} & \textbf{Value} \\ \midrule
Training steps & 300,000 \\
Warm up steps & 375 \\
Batch size & 512 \\
Optimizer & Adam \\
Learning rate & 6e-4 \\
Learning Rate Decay & Linear \\
Adam $\epsilon$ & 1e-6 \\
Adam $\beta$ & (0.9, 0.98) \\
Weight decay & 0.01 \\
Non-causal percent & 0.25 \\
\bottomrule
\end{tabular}
\captionsetup{type=table}
\caption{Optimization hyperparameters for language-only pretraining.}
\label{tbl:hyperparam:lang:pt:opt}
\end{minipage}

\subsection{Multitask Finetuning and Instruction Tuning}
\label{app:hyperparam:lang:multi}

We provide the detailed settings of language-only multitask finetuning and instruction tuning with \ours{} in Table~\ref{tbl:hyperparam:lang:multi}.

\begin{table}[ht]
\centering
\small
\renewcommand\tabcolsep{3.5pt}
\begin{tabular}{lcc}
\toprule
\textbf{Hyperparameters} & \textbf{Multitask Finetuning} & \textbf{Instruction Tuning} \\ \midrule
Training steps & 20,000 & 30,000 \\
Warm up steps & 2,000 & 3,000 \\
Batch size & 256 & 512\\
Optimizer & Adam & Adam \\
Learning rate & 1e-4 & 1e-4 \\
Adam $\epsilon$ & 1e-6 & 1e-6 \\
Adam $\beta$ & (0.9, 0.98) & (0.9, 0.98) \\
Weight decay & 0.01 & 0.01 \\
Max length & 2048 & 1024 \\
Dropout of non-causal model & 0.1 & 0.1 \\
Dropout of causal model & 0.0 & 0.0 \\
\bottomrule
\end{tabular}
\caption{Hyperparameters used for language-only multitask finetuning and instruction tuning. 
}
\label{tbl:hyperparam:lang:multi}
\end{table}

\section{Datasets Used for Language-Only Experiments}
\label{app:corpora:pt:lang}

\subsection{Pretraining}
\label{app:corpora:data:lang:pt}

Language-only \ours{} is pretrained on Pile~\citep{pile}, which is an 800 GB English text corpus combining 22 diverse sources.
We exclude data sources of GitHub, arXiv and PubMed Central from the original Pile.
Thus the pretraining corpus we used is composed of 19 sources, divided into the following five categories:
\begin{itemize}[leftmargin=*]
\item \textbf{Academic}: FreeLaw, USPTO Backgrounds, PhilPapers, NIH Exporter, PubMed Abstracts
\item \textbf{Internet}: Pile-CC, OpenWebText2, StackExchange, Wikipedia (English)
\item \textbf{Prose}: BookCorpus2, Books3, Gutenberg~\citep[PG-19]{pg19}
\item \textbf{Dialogue}: OpenSubtitles~\citep{opensubtitle}, Youtube Subtitles, EuroParl~\citep{europarl}, Hacker News, Ubuntu IRC
\item \textbf{Miscellaneous}: Enron Emails~\citep{enron}, DM Mathematics~\citep{dmmath}
\end{itemize}

\subsection{Multitask Finetuning and Instruction Tuning}
\label{app:corpora:data:lang:ft}
We list the datasets we used for language-only multitask finetuning and instruction tuning.
\begin{itemize}[leftmargin=*]
\item \textbf{Natural Language Inference} is to determine whether a hypothesis is true (entailment), false (contradiction) or undetermined (neutral) given a premise.
We use the following datasets: ANLI~\citep{anli}, CB~\citep{cb}, MNLI~\citep{mnli}, QNLI~\citep{squad}, RTE~\citep{rte1,rte2,rte3,rte5}, SNLI~\citep{snli} and WNLI~\citep{wsc}.
\item \textbf{Sentiment Classification} is to determine the emotional tone of a piece of text, whether it is positive or negative: IMDB~\citep{imdb}, SST-2~\citep{sst2}, Sentiment140~\citep{sent140}, Yelp~\citep{yelp}.
\item \textbf{Paraphrase Detection} is to detect the semantic similarity of two sentences: QQP~\citep{glue}, MRPC~\citep{mrpc}, Paws Wiki~\citep{paws}.
\item \textbf{Coreference Resolution} is to determine if two expressions refer to the same entity in a text: DPR~\citep{dpr}, Winogrande~\citep{winogrande}, WSC~\citep{wsc}.
\item \textbf{Commonsense Reasoning} evaluates the ability to perform physical or social commonsense: HellaSwag~\citep{hellaswag}, PiQA~\citep{piqa}, COPA~\citep{copa}.
\item \textbf{Reading Comprehension} is to answer some questions conditioned on a given passage: DROP~\citep{drop}, SQuADv1~\citep{squad}, SQuADv2~\citep{squad2}, OBQA~\citep{obqa}, BoolQ~\citep{boolq}.
\item \textbf{Miscellaneous} consists of some additional datasets: CoLA~\citep{cola}, WiC~\citep{wic}, TREC~\citep{trec1,trec2}.
\item \textbf{Closed-Book QA} is to answer a question without external knowledge: ARC-easy~\citep{arc}, NQ~\citep{nq1,nq2}.
\item \textbf{Struct to Text} is to construct a natural language description for some structured data: CommonGen~\citep{commongen}, E2ENLG~\citep{e2enlg}.
\item \textbf{Summarization} is to generate a summary of a given passage: AESLC~\citep{aeslc}, SamSum~\citep{samsum}, XSum~\citep{xsum}.
\end{itemize}
Furthermore, we utilize the hand-crafted templates from FLAN~\citep{flan}, which composes ten templates for each dataset.
For multitask finetuning, we apply only the first template for each dataset. 
For instruction tuning, we apply all the ten templates.

\subsection{In-Context Learning}
\label{app:corpora:data:lang:fewshot}
We conduct experiments of in-context learning on four categories:
\begin{itemize}[leftmargin=*]
\item Cloze and completion tasks: StoryCloze~\citep{storycloze}, HellaSwag~\citep{hellaswag}
\item Winograd-style tasks: Winograd~\citep{wsc}, Winogrande~\citep{winogrande}
\item Commonsense reasoning: ARC-easy/ARC-challenge~\citep{arc}, PIQA~\citep{piqa}
\item Two datasets from SuperGLUE benchmark~\citep{superglue}: BoolQ~\citep{boolq}, Copa~\citep{copa}
\end{itemize}

\section{Detailed Results of Multitask Finetuning in Section~\ref{sec:lang:multitask}}
\label{app:results:lang:multitask}

We list the full results of language-only multitask finetuning for all task clusters in our experiments.
Results of natural language inference are shown in Table~\ref{tbl:multi:nli}.
Results of sentiment classification are shown in Table~\ref{tbl:multi:sent}.
Results of paraphrase detection are shown in Table~\ref{tbl:multi:para}.
Results of reading comprehension are shown in Table~\ref{tbl:multi:reading}.
Results of coreference resolution are shown in Table~\ref{tbl:multi:coref}.
Results of miscellaneous cluster are shown in Table~\ref{tbl:multi:misc}.
Results of commonsense reasoning are shown in Table~\ref{tbl:multi:common}.
Results of struct to text are shown in Table~\ref{tbl:multi:struct}.
Results of closed-book QA are shown in Table~\ref{tbl:multi:closeqa}.
Results of text summarization are shown in Table~\ref{tbl:multi:summ}.

\begin{table}[tp!]
\centering
\renewcommand\tabcolsep{3.5pt}
\begin{tabular}{l c c c c c c c c c}
\toprule
 & ANLI R1 & ANLI R2 & ANLI R3 & CB & MNLI-m & QNLI & RTE & SNLI & WNLI  \\
 \midrule
GPT & 52.9 & 45.5 & 43.5 & 89.3 & 81.1 & 90.1 & 79.4 & 85.3 & 18.3 \\
\ours{} & \textbf{72.6}& \textbf{54.1}& \textbf{49.5}& \textbf{91.1}& \textbf{88.9}& \textbf{93.8}& \textbf{87.7}& \textbf{90.0}& \textbf{84.5} \\
\bottomrule
\end{tabular}
\caption{Multitask finetuning results of natural language inference.}
\label{tbl:multi:nli}
\end{table}

\begin{table}[tp!]
\centering
\begin{tabular}{l c c c c c c c c c}
\toprule
 & IMDB & SST-2 & Sent140 & Yelp \\
\midrule
GPT & 95.7 & 93.5 & 85.2 & 97.2 \\
\ours{} & \textbf{96.5}& \textbf{94.7}& \textbf{89.1}& \textbf{97.9} \\
\bottomrule
\end{tabular}
\caption{Multitask finetuning results of sentiment classification.}
\label{tbl:multi:sent}
\end{table}

\begin{table}[tp!]
\centering
\begin{tabular}{l c c c c c c c c c}
\toprule
 & QQP & MRPC & PAWS Wiki \\
\midrule
GPT & 84.4 & 78.9 & 88.4 \\
\ours{} & \textbf{88.0}& \textbf{86.8}& \textbf{94.1} \\
\bottomrule
\end{tabular}
\caption{Multitask finetuning results of paraphrase detection.}
\label{tbl:multi:para}
\end{table}

\begin{table}[tp!]
\centering
\begin{tabular}{l c c c c c c c c c}
\toprule
 & DROP & SQuADv1 & SQuADv2 & OBQA & BoolQ \\
 Metric & f1 & f1 & f1 & acc & acc \\
\midrule
GPT & 38.5 & 84.0 & 72.0 & 49.6 & 78.4 \\
\ours{} & \textbf{45.2}& \textbf{89.3}& \textbf{84.1}& \textbf{61.8}& \textbf{85.0} \\
\bottomrule
\end{tabular}
\caption{Multitask finetuning results of reading comprehension.}
\label{tbl:multi:reading}
\end{table}

\begin{table}[tp!]
\centering
\begin{tabular}{l c c c c c c c c c}
\toprule
 & DPR & Winogrande \\
\midrule
GPT & 71.6 & 62.5 \\
\ours{} & \textbf{87.8}& \textbf{80.8} \\
\bottomrule
\end{tabular}
\caption{Multitask finetuning results of coreference resolution.}
\label{tbl:multi:coref}
\end{table}

\begin{table}[tp!]
\centering
\begin{tabular}{l c c c c c c c c c}
\toprule
 & CoLA & WIC & TREC \\
\midrule
GPT & 79.7 & 63.9 & 97.2 \\
\ours{} & \textbf{87.3}& \textbf{67.6}& \textbf{98.0} \\
\bottomrule
\end{tabular}
\caption{Multitask finetuning results of miscellaneous cluster.}
\label{tbl:multi:misc}
\end{table}

\begin{table}[tp!]
\centering
\begin{tabular}{l c c c c c c c c c}
\toprule
 & HellaSwag & PiQA & CoPA \\
\midrule
GPT & 56.5 & 67.3 & 66.0 \\
\ours{} & \textbf{83.2}& \textbf{76.4}& \textbf{93.0} \\
\bottomrule
\end{tabular}
\caption{Multitask finetuning results of commonsense reasoning.}
\label{tbl:multi:common}
\end{table}

\begin{table}[tp!]
\centering
\begin{tabular}{l c c c c c c c c c}
\toprule
 & NQ & ARC-e \\
\midrule
GPT & 15.3 & 61.1 \\
\ours{} & \textbf{16.5}& \textbf{72.1} \\
\bottomrule
\end{tabular}
\caption{Multitask finetuning results of closed-book question answering.}
\label{tbl:multi:struct}
\end{table}

\begin{table}[tp!]
\centering
\begin{tabular}{l c c c c c c}
\toprule
 & \multicolumn{3}{c}{CommonGen} & \multicolumn{3}{c}{E2ENLG} \\
 \cmidrule(r){2-4} \cmidrule(l){5-7}
 & {Rouge-1} & {Rouge-2} & {Rouge-L} & {Rouge-1} & {Rouge-2} & {Rouge-L} \\
 \midrule
GPT & \textbf{47.3} & 18.7 & \textbf{41.0} & 64.2 & \textbf{37.0} & \textbf{47.5} \\
\ours{} & 46.8& \textbf{18.8}& 40.7& \textbf{64.4}& 36.4& \textbf{47.5} \\
\bottomrule
\end{tabular}
\caption{Multitask finetuning results of struct to text.}
\label{tbl:multi:closeqa}
\end{table}

\begin{table}[tp!]
\centering
\renewcommand\tabcolsep{3.5pt}
\footnotesize
\begin{tabular}{l c c c c c c c c c}
\toprule
 & \multicolumn{3}{c}{AESLC} & \multicolumn{3}{c}{SamSum} & \multicolumn{3}{c}{XSum} \\
 \cmidrule(r){2-4} \cmidrule(l){5-7} \cmidrule(l){8-10}
 & {Rouge-1} & {Rouge-2} & {Rouge-L} & {Rouge-1} & {Rouge-2} & {Rouge-L} & {Rouge-1} & {Rouge-2} & {Rouge-L} \\
 \midrule
GPT & 29.1 & \textbf{15.3} & 28.7 & 45.3 & 20.2 & 36.7 & 29.8 & 10.5 & 24.0 \\
\ours{} & \textbf{30.3}& 15.0& \textbf{29.5}& \textbf{46.5}& \textbf{21.7}& \textbf{38.2}& \textbf{31.3}& \textbf{11.6}& \textbf{25.3} \\
\bottomrule
\end{tabular}
\caption{Multitask finetuning results of text summarization.}
\label{tbl:multi:summ}
\end{table}

\section{Hyperparameters of Vision-Language Experiments}

\subsection{Hyperparameters of Vision-Language Pretraining}
\label{app:hyperparam:pt:vl}

We report the detailed pretraining hyperparameter settings of the vision-language \ours{} in Table~\ref{tbl:hyperparam:vl:pt:model} and report the optimization hyperparameters in Table~\ref{tab:vl:hyperparam:pt:opt}.

\noindent\hspace{0.03\linewidth}\begin{minipage}[c]{0.5\textwidth}
\small
\centering
\begin{tabular}{lcc}
\toprule
\textbf{Hyperparameters} & \textbf{Non-causal} & \textbf{Semi-causal} \\ \midrule
Number of layers & 12 & 24 \\
Hidden size & 768 & 1,024 \\
FFN inner hidden size & 3,072 & 4,096 \\
Attention heads & 12 & 16 \\
Dropout & 0.1 & 0.2 \\
Attention Dropout & 0.1 & 0.2 \\
Vocabulary Size & 50,259 & 50,259 \\
Pretraining length & 236 & 1024 \\
Freeze in pretraining & No & No \\
Position Embedding & Learnable & Sinusoidal \\
\bottomrule
\end{tabular}
\captionsetup{type=table}
\caption{Hyperparameters of non-causal and semi-causal models for vision-language pretraining.}
\label{tbl:hyperparam:vl:pt:model}
\end{minipage}
\hfill
\begin{minipage}[c]{0.4\textwidth}
\small
\centering
\begin{tabular}{lr}
\toprule
\textbf{Hyperparameters} & \textbf{Value} \\ \midrule
Training steps & 350,000 \\
Warm up steps & 2,500 \\
Batch size & 256 \\
Optimizer & AdamW \\
Learning rate & 1e-4 \\
Learning Rate Decay & Linear \\
Adam $\epsilon$ & 1e-8 \\
Adam $\beta$ & (0.9, 0.98) \\
Weight decay & 0.01 \\
Image size  & 224x224 \\
Patch size  & 16 \\
\bottomrule
\end{tabular}
\captionsetup{type=table}
\caption{Optimization hyperparameters for vision-language pretraining.}
\label{tab:vl:hyperparam:pt:opt}
\end{minipage}

\subsection{Hyperparameters in Vision-Language Finetuning}
\label{app:hyperparam:ft:vl}
We report the finetuning settings along with the prompts in Table~\ref{tab:vl:ft-hparam}. The vision-language \ours{} applies a 384x384 image size and greedy decoding for all finetuning tasks.

\begin{table}[ht]
\centering
\small
\renewcommand\tabcolsep{3.5pt}
\begin{tabular}{lcccc}
\toprule
\textbf{Task} & \textbf{Learning rate} & \textbf{Batch size} & \textbf{Steps} & \textbf{Language prompt} \\ \midrule
VQAv2 & 1e-5 & 64 & 140k & question: [question] answer: [answer] \\
VQA Karpathy & 1e-5 & 64 & 140k & question: [question] answer: [answer] \\
OK-VQA & 1e-5 & 8 & 10k & question: [question] answer: [answer] \\
COCO Caption & 1e-5 & 64 & 100k & caption: [caption] \\
NLVR$^2$ & 1e-5 & 32 & 54k & it is [label] \\
E-SNLI-VE label & 1e-5 & 64 & 54k & it is [label] because [explanation] \\
E-SNLI-VE explanation & 1e-5 & 64 & 54k & it is [label] because [explanation] \\
\bottomrule
\end{tabular}
\caption{Summary of hyperparameters for vision-language finetuning.}
\label{tab:vl:ft-hparam}
\end{table}

\end{document}